\documentclass[journal]{IEEEtran}
\ifCLASSINFOpdf
\else
\fi
\usepackage{graphicx}
\usepackage[misc]{ifsym}
\usepackage[linesnumbered,ruled,vlined]{algorithm2e}
\usepackage{xcolor}
\usepackage[utf8]{inputenc} 
\usepackage[T1]{fontenc}    
\usepackage{hyperref}       
\usepackage{url}            
\usepackage{booktabs}       
\usepackage{amsfonts}       
\usepackage{nicefrac}       
\usepackage{microtype}      
\usepackage{blindtext}
\usepackage[linesnumbered,ruled,vlined]{algorithm2e}
\usepackage{amssymb}
\usepackage{amsmath} 
\usepackage[T1]{fontenc}
\usepackage{lmodern}
\usepackage{soul}
\usepackage{makecell}
\soulregister\cite7
\soulregister\ref7
\soulregister\eqref7

\newtheorem{definition}{Definition}
\newtheorem{theorem}{Theorem}
\newtheorem{lemma}{Lemma}
\newtheorem{assumption}{Assumption}
\newtheorem{corollary}{Corollary}

\begin{document}
%
\title{FastAdaBelief: Improving Convergence Rate for Belief-based Adaptive Optimizers by Exploiting Strong Convexity}


\author{\IEEEauthorblockN{Yangfan Zhou, Kaizhu Huang, Cheng Cheng, Xuguang Wang, Amir Hussain, and Xin Liu*}
\thanks{Y.Zhou and X. Liu are with School of Nano-Tech and Nano-Bionics, University of Science and Technology of China, 96 Jinzhai Road, Hefei City, Anhui Province, 230026 China.}
\thanks{Y. Zhou, C. Cheng, X. Wang, and X. Liu are with Suzhou Institute of Nano-Tech and Nano-Bionics (SINANO), Chinese Academy of Sciences, 398 Ruoshui Road, Suzhou Industrial Park, Suzhou City, Jiangsu Province, 215123 China.}
\thanks{K. Huang is with Data Science Research Center, Duke Kunshan University, No. 8 Duke Avenue, Kunshan, 215316, China.}
\thanks{A. Hussain is with the School of Computing, Edinburgh Napier University, Edinburgh, EH11 4BN, UK.}
\thanks{C. Cheng is also with Gusu Laboratory of Materials, 388 Ruoshui Road, Suzhou, Jiangsu Province, 215123 China.}
\thanks{E-mail addresses: yfzhou2020@sinano.ac.cn, kaizhu.huang@dukekunshan.edu.cn, ccheng2017@sinano.ac.cn,  xgwang2009@sinano.ac.cn, A.Hussain@napier.ac.uk, xliu2018@sinano.ac.cn.}
\thanks{* Corresponding author: Xin Liu (email: xliu2018@sinano.ac.cn).}}
\markboth{Journal of \LaTeX\ Class Files,~Vol.~00, No.~0, Nov.~2021}%
{Shell \MakeLowercase{\textit{et al.}}}
%



\IEEEtitleabstractindextext{%
\begin{abstract}
AdaBelief, one of the current best optimizers, demonstrates superior generalization ability over the popular Adam algorithm by viewing the exponential moving average of observed gradients. AdaBelief is theoretically appealing in that it has a data-dependent $O(\sqrt{T})$ regret bound when objective functions are convex, where $T$ is a time horizon. It remains however an open problem whether the convergence rate  can be further improved without sacrificing its generalization ability.
To this end,  we make a first attempt in this work and design a novel optimization algorithm called FastAdaBelief that aims to exploit its strong convexity in order to achieve an even faster convergence rate. In particular, by adjusting the step size that better considers strong convexity and prevents fluctuation, our proposed FastAdaBelief demonstrates excellent generalization ability as well as superior convergence. As an important theoretical contribution, we prove that FastAdaBelief attains a data-dependant $O(\log T)$ regret bound, which is substantially lower than AdaBelief in strongly convex cases. On the empirical side, we validate  our theoretical analysis with extensive experiments in scenarios of strong convexity and non-convexity using three popular baseline models. Experimental results are very encouraging: FastAdaBelief converges the quickest  in comparison to all mainstream algorithms while maintaining an excellent generalization ability, in cases of both strong convexity or non-convexity. FastAdaBelief is thus posited as a new benchmark model for the research community.
\end{abstract}

}

\maketitle

\IEEEdisplaynontitleabstractindextext

%
\IEEEpeerreviewmaketitle

\section{Introduction}\label{introduction}
\IEEEPARstart{T}{he} training process is a significant stage in many fields of artificial neural networks such as deep learning \cite{Huang2020}, transfer learning \cite{Jiang2017} and meta learning \cite{Wang2021}. From an optimization perspective, the purpose of the training process is to minimize (or maximize) the loss value (or reward value), and thus can be  considered as an optimization process~\cite{Jin2019}.  As a popular paradigm,  the training process can be conducted in a supervised way that requires a large number of labeled samples in order to achieve satisfactory performance~\cite{Jia2021}.  Whilst on the one hand, this can be very difficult to deploy in practice due to the high cost involved in  annotating samples manually (or even automatically)~\cite{Niu2020}; on the other hand,  even with sufficient labelled data, it is still a formidable challenge on how to design both fast and accurate training or optimization algorithms. To tackle this problem, many researchers have made attempts to improve the convergence speed of optimization algorithms, so as to both reduce the need  for labeled samples and speed up  the training process with available data at hand~\cite{Khan2018,Mukkamala2017}. Specifically, online learning is often used to accomplish such training tasks since it does not require information to be collected in batches at the same time \cite{Zhou2020}.

\begin{table*}
  \caption{Comparison of performance on convergence and generalization ability of FastAdaBelief and the current mainstream optimizers.}
  \label{table1}
  \centering
  \begin{tabular}{lclcc}
    \toprule
    \textbf{Optimizer}    &\textbf{Loss Function}     &\textbf{Regret Bound} &\textbf{Convergence}   &\textbf{Generalization} \\
    \midrule
    SGD (\cite{Zinkevich2003})         &\emph{convex}            &$O(\sqrt{T})$   &\emph{slow}       &\color{blue}{\emph{excellent}}     \\
    Adam(\cite{Kingma2015})         &\emph{convex}            &$O(\sqrt{T})$   &\emph{medium}     &\emph{poor}    \\
    SAdam (\cite{Wang2020})        &\emph{strongly convex}   &$O(\log T)$    &\color{blue}{\emph{fast}}       &\emph{poor}     \\
    AdaBelief (\cite{Zhuang2020})    &\emph{convex}            &$O(\sqrt{T})$   &\emph{medium}     &\color{blue}{\emph{excellent}}     \\
    FastAdaBelief (Ours)  &\emph{strongly convex}   &$O(\log T)$    &\color{blue}{\emph{fast}}       &\color{blue}{\emph{excellent}}    \\
    \bottomrule
  \end{tabular}
\end{table*}

One  classic online optimization algorithm is online Stochastic Gradient Descent (SGD)~\cite{Zinkevich2003}. SGD has been extensively applied over the last few decades in many training tasks of deep learning owing to its simple logic and good generalization ability~\cite{Lei2020,Gu2021}. However, SGD suffers from the limitation of slow convergence.  This disadvantage hinders its application especially in  large-scale problems which may take extremely long to converge. To address this issue, researchers have developed various methods to speed up the SGD convergence rate. For example, one type of methods focus on exploring first-order momentum to accelerate SGD; such methods include SGD with momentum~\cite{Sutskever2013} and Nesterov momentum~\cite{Nesterov1983}. Typically adopting a fixed step size, these methods may not be conducive to accelerate the convergence rate. To alleviate this problem,
recent studies including the popular Adam~\cite{Kingma2015} and AMSGrad approaches \cite{Reddi2018},  attempt to apply second-order momentum and prefer an adaptive step size while maintaining the first-order momentum.

As one of the most successful adaptive online algorithms, Adam enjoys a fast convergence which is guaranteed with the regret bound of $O(\sqrt{T})$. Despite its outstanding performance, Reddi \emph{et al.}  indicated that Adam has the issue of non-convergence~\cite{Reddi2018}, which is caused by not satisfying $\Gamma_t\succeq0$ for all $t\in\{1,\ldots,T\}$, where $\Gamma_t = \frac{\sqrt{\mathbf{v}_t}}{\alpha_t} - \frac{\sqrt{\mathbf{v}_{t-1}}}{\alpha_{t-1}}$. Moreover, another limitation with Adam is that it can lead to poorer generalization ability compared to SGD. To tackle this issue, many variants of Adam have been further proposed. For instance, Luo \emph{et al.} \cite{Luo2019} proposed AdaBound with a dynamic bound on the learning rate; Zaheer \emph{et al.} considered the effect of increasing mini-batch size, and proposed Yogi \cite{Zaheer2018}; Liu \emph{et al.} developed RAdam \cite{Liu2020} to rectify the variance of the learning rate; Balles and Hennig dissected Adam in the sign, magnitude, and variance of stochastic gradients, and proposed MSVAG \cite{Balles2018}; Loshchilov and Hutter proposed AdamW~ \cite{Loshchilov2019} to decouple the weight decay from the loss function.

Although these variants perform better than Adam in generalization ability, there is still a generalization gap compared with SGD on large-scale datasets. To fill this gap, Zhuang \emph{et al.} proposed AdaBelief \cite{Zhuang2020}, which adapts the step size by the belief in observed gradients and leads to superior generalization compared to Adam. Specifically, AdaBelief re-designs the second-order momentum into a novel form that is closer to the ideal choice. Moreover, the regret bound of AdaBelief is proved to be $O(\sqrt{T})$ when loss functions are convex.

Albeit its success, it however remains an open problem \emph{if the convergence rate of AdaBelief can be further improved without sacrificing its generalization ability in certain cases}. To this end,  in this work we  design a novel optimization algorithm called FastAdaBelief that aims to exploit the strong convexity in order to achieve an even faster convergence rate whilst maintaining an excellent generalization ability.  In particular, by adjusting the step size that better considers strong convexity, appropriately utilizes curvature information, and prevents fluctuation,  our proposed FastAdaBelief  attains
a substantially  lower data-dependant regret bound, which
generally promotes AdaBelief from a sublinear level $O(\sqrt T)$ to a logarithmic level $O\big(\sum_{i=1}^n\log\big(\left\|g_{1:T,i}\right\|^2\big)\big)$, and to $O(\log T)$ in the worst case in the strongly convex scenarios. Despite this, SGD and some other first-order optimization algorithms can  achieve a data-independent bound $O(\log T)$ for online strongly convex optimization, and the data-dependent regret bound $O\big(\sum_{i=1}^n\log\big(\left\|g_{1:T,i}\right\|^2\big)\big)$ can be much tighter than the data-independent bound whenever the gradients are sparse or small such that $\left\|g_{1:T,i}\right\|^2\ll TG_{\infty}^2$. To the best of our knowledge, FastAdaBelief is a first attempt at designing a powerful optimizer that converges faster with a logarithmic regret bound while simultaneously maintaining an excellent generalization ability.

Note that  Wang \emph{et al.} proposed SAdam~\cite{Wang2020} to implement Adam into strong convexity, which is also able to accelerate the regret bound of Adam from $O(\sqrt{T})$ to $O(\log T)$. However, SAdam generally has a poor generalization ability as it is rooted from Adam, which may hence limit its application in practice.  On closer examination, FastAdaBelief adopts the new second order form that is significantly different  from the form of SAdam; this posits a new challenge for the convergence analysis of FastAdaBelief. Additionally, in order to fit strongly convex conditions, FastAdaBelief  designs a tailored diagonal matrix of the second order momentum, which also lead to a non-trivial challenge in  the  convergence analysis compared to Adam.  In summary, the performance based on convergence and generalization ability of FastAdaBelief and the current mainstream optimizers can be seen in Table~\ref{table1}.

Our major contributions are summarized below:
\begin{itemize}
  \item We propose a fast variant of AdaBelief, named FastAdaBelief, to further improve the convergence rate under strongly convex conditions. We show that FastAdaBelief can lead to an adaptive stepsize that is more in line with an ideal optimizer.
  \item We provide a convergence analysis for FastAdaBelief that presents a data-dependant $O\big(\sum_{i=1}^n\log\big(\left\|g_{1:T,i}\right\|^2\big)\big)$ guaranteed regret bound, which is substantially better than AdaBelief.
  \item We conduct extensive experiments to demonstrate that FastAdaBelief outperforms other state-of-the-art main-stream optimization algorithms in a variety of tasks. Interestingly, even in case of non-convexity, FastAdaBelief shows consistently superior performance over other state-of-the-art algorithms with all benchmark datasets.

\end{itemize}

\section{Notation and Preliminaries} \label{related}
\subsection{Notation}
Since this paper uses a lot of symbols, for brevity, we summarize the notations in Table \ref{notation} below.

\begin{table}[h]
  \caption{The summary of notations.}
  \label{notation}
  \centering
  \begin{tabular}{l|l}
    \toprule
    \midrule
    \textbf{Symbol}  &\textbf{Meaning}\\
    \midrule
    \toprule
    $\mathbf{x}$  &lowercase bold letters represent vectors\\
    \midrule
    $\mathbf{x}_t$  &the value of vector $\mathbf{x}$ at time $t$\\
    \midrule
    $x_{t,i}$  &the $i$-th coordinate of vector $\mathbf{x}_t$\\
    \midrule
    $M$  &capital letters represent matrices\\
    \midrule
    $\mathcal{M}_{+}^n$  &\makecell[l]{the set of $n$ dimensional positive definite \\ matrices}\\
    \midrule
    $\|\cdot\|$  &the $\ell_2$-norm\\
    \midrule
    $\|\cdot\|_{\infty}$  &the $\ell_{\infty}$-norm\\
    \midrule
    $\|\mathbf{x}\|_{M}^2=\mathbf{x}^{\mathsf{T}}M\mathbf{x}$  &the $M$-weighted $\ell_2$-norm\\
    \midrule
    $f_t(\cdot)$  &the loss function at time $t$\\
    \midrule
    $\mathbf{g}_t$  &the gradient of the loss function $f_t(\cdot)$\\
    \midrule
    $\mathbf{g}_{1:T,i}=[g_{1,i},\ldots,g_{T,i}]$  &\makecell[l]{the sequence consisting of \\ the $i$-th element of \\ the gradient sequence $\{\mathbf{g}_1,\ldots,\mathbf{g}_{T}\}$}\\
    \midrule
    \makecell[l]{$\prod_{\mathcal{F}, M}(\mathbf{x}) =$ \\ \\ $\arg\min_{\mathbf{y}\in\mathcal{F}}\|\mathbf{y}-\mathbf{x}\|_M^2$}  &\makecell[l]{the $M$-weighted projection operation of \\ $\mathbf{x}$ on $\mathcal{F}$, where $\mathbf{x}\in\mathbb{R}^n$}\\
    \midrule
    $\mathbf{x}^2$  &\makecell[l]{the element-wise square, \\ where $\mathbf{x}\in\mathbb{R}^n$}\\
    \midrule
    $\frac{\mathbf{x}}{\mathbf{y}}$  &\makecell[l]{the element-wise division, \\ where $\mathbf{x},\mathbf{y}\in\mathbb{R}^n$}\\
    \midrule
    $\sqrt{\mathbf{x}}$  &\makecell[l]{the element-wise square root, \\ where $\mathbf{x}\in\mathbb{R}^n$}\\
    \midrule
    $A=\mathrm{diag}\{\mathbf{x}\}$  &\makecell[l]{the fact that $A$ is a diagonal matrix \\ composed of the elements of vector $\mathbf{x}$}\\
    \midrule
    $I_d$  & a $d\times d$ identity matrix\\
    \midrule
    $\mathbf{x}^*$  &\makecell[l]{the best decision in hindsight,\\ i.e., $\mathbf{x}^* = \min_{\mathbf{x}\in\mathcal{F}}\sum_{t=1}^T f_t(\mathbf{x})$}\\
    \midrule
    \bottomrule
  \end{tabular}
\end{table}

\subsection{Online Learning}
Machine learning (ML) plays an important role in the field of artificial intelligence. Moreover, offline learning in ML is usually expected to enable a batch of tasks at the same time, but this situation is difficult to meet. In contrast, online learning based on regret considers a sequential setting in which tasks are revealed one by one. Online learning can better adapt to complex and changeable practical applications and  has become a prominent paradigm for machine learning, which is attractive in both theory and practice~\cite{ShalevShwartz2011}. Within this paradigm, a learner  iteratively generates  a decision $\mathbf{x}_t$ from a convex and compact domain $\mathcal{F}\subset\mathbb{R}^n$ in each round $t\in\{1,\ldots,T\}$. In response, an adversary produces a convex loss function $f_t(\cdot):\mathcal{F}\rightarrow\mathbb{R}$ in round $t$, which causes the learner to suffer the loss $f_t(\mathbf{x}_t)$. The goal of the learner is to generate a decision $\mathbf{x}_t$ so that the regret can decrease quickly as $T$. Moreover, the regret is defined as follows:
\begin{equation}\label{eq1}
    R(T) = \sum_{t=1}^T f_t(\mathbf{x}_t) - \min_{\mathbf{x}\in\mathcal{F}}\sum_{t=1}^T f_t(\mathbf{x}).
\end{equation}

To improve the generalization ability of the Adam optimizer family, AdaBelief fully considers the curvature information of loss functions, which will be introduced  as part of preliminary background in the next subsection.

\subsection{AdaBelief}
The algorithm design of AdaBelief is shown in Algorithm~\ref{AdaBelief}. Reviewing that Adam designs its second-order momentum as the following form:
\begin{equation*}\label{0-1}
 \mathbf{v}_t = \beta_2\mathbf{v}_{t-1} + (1-\beta_2)\mathbf{g}_t^2,
\end{equation*}
where $\mathbf{g}_t$ is the gradient, and $t\in\{1,\ldots,T\}$. Moreover, Algorithm~\ref{AdaBelief} shows that AdaBelief's novel seconde-order momentum is designed as:
\begin{equation*}\label{0-2}
 \mathbf{s}_t = \beta_2\mathbf{s}_{t-1} + (1-\beta_2)(\mathbf{g}_t-\mathbf{m}_t)^2,
\end{equation*}
where $\mathbf{m}_t$ is the first-order momentum. Note that since the second order momentums of AdaBelief and Adam are quite different, thereby that of Adabelief is symbolized by $\mathbf{s}_t$, and that of Adam by $\mathbf{v}_t$~\cite{Kingma2015}.

\begin{figure}
    \centering
		\includegraphics[scale=.35]{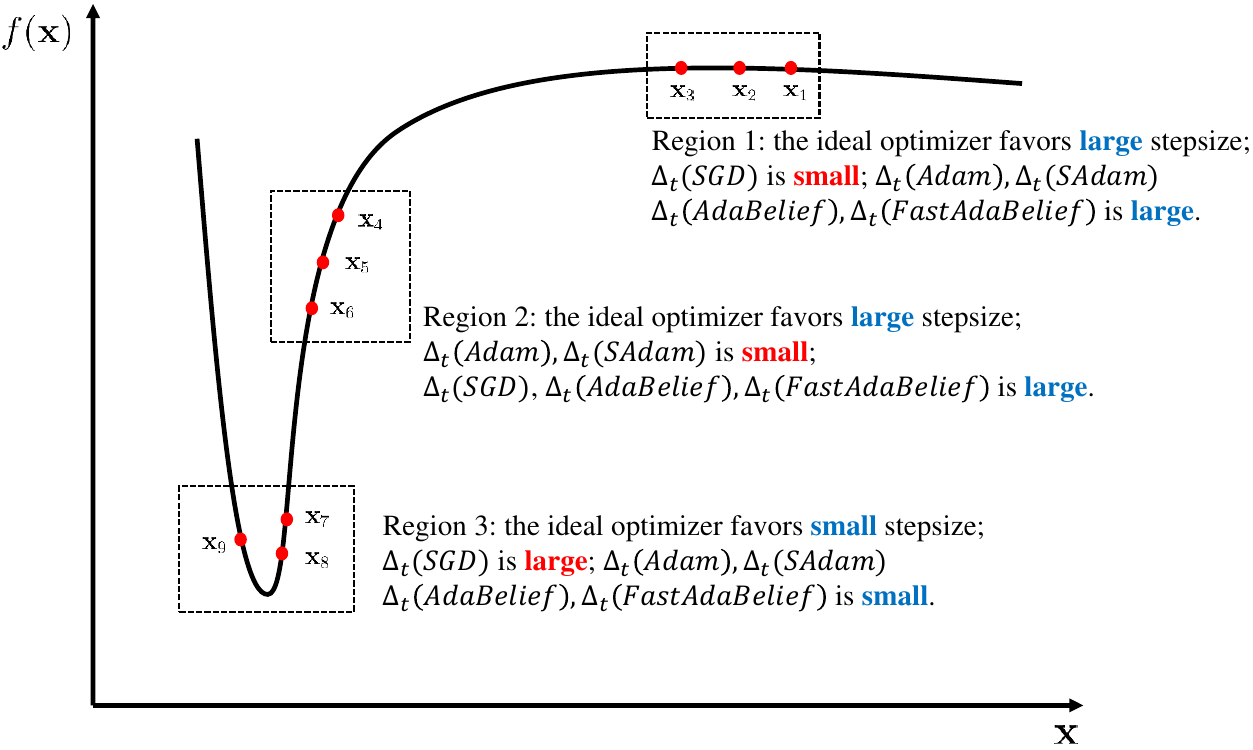}
	    \caption{An ideal optimizer considers the curvature of the loss function and prefers adaptive stepsize. $\Delta_t$ denotes the stepsize. FastAdaBelief selects a stepsize  more in line with the ideal optimizer (see more details in Table~\ref{tableStepsize})(The figure is adapted from \cite{Zhuang2020}).}
	    \label{fig0}
\end{figure}

\begin{algorithm}
\DontPrintSemicolon 
\KwIn{$ \beta_{1} , \beta_{2}$}
\KwOut{$\mathbf{x}_{t+1}$}
\textbf{Initialize:} $\mathbf{x}_0, \mathbf{m}_0, \mathbf{s}_0$\\
\For{$t = 1 \ldots T$} {
  $t \gets t + 1$  \\
  $\alpha_t \gets \frac{\alpha}{\sqrt{t}}$ \\
  $\mathbf{g}_t \gets \nabla f_t(\mathbf{x}_t)$ \\
  $\mathbf{m}_t \gets \beta_{1}\mathbf{m}_{t-1} + (1-\beta_{1})\mathbf{g}_t$ \\
  $\mathbf{s}_t \gets \beta_{2}\mathbf{s}_{t-1} + (1-\beta_{2})(\mathbf{g}_t - \mathbf{m}_t)^2$ \\
  $\hat{\mathbf{s}}_{t} \gets \max\{\hat{\mathbf{s}}_{t-1}, \mathbf{s}_{t}\}$ \\
  $\hat{S}_t \gets \mathrm{diag}\{\hat{\mathbf{s}}_{t}\}$ \\
  $\mathbf{x}_{t+1} \gets \prod_{\mathcal{F},\sqrt{\hat{S}_t}}\left(\mathbf{x}_{t} - \frac{\alpha_t\mathbf{m}_t}{\sqrt{\hat{S}_t}+\epsilon} \right)$
}
\Return{$\mathbf{x}_{t+1}$}\;
\caption{AdaBelief}
\label{AdaBelief}
\end{algorithm}

To better illustrate the difference of various optimizers, we exploit one illustrative example  similarly utilized by AdaBelief as shown in Figure~\ref{fig0}. In region 1 where the loss function is flat,  the gradient $\mathbf{g}_t$ and $|\mathbf{g}_t(\mathbf{x_1}) - \mathbf{g}_t(\mathbf{x_2})|$ are both very small. In fact, a large stepsize should be taken in this case for the efficiency of the optimizer. In this case, AdaBelief and Adam both take large stepsizes, but SGD takes a small one.

In region 2 called the ``large gradient, small curvature" case, the gradient $\mathbf{g}_t$ and $\mathbf{v}_t$ are both large while $|\mathbf{g}_t - \mathbf{g}_{t-1}|$ and $\mathbf{s}_t$ are both small. In fact, the stepsize of an ideal optimizer should be increased in this region. To this end, AdaBelief takes a large stepsize since its denominator $\sqrt{\mathbf{s}_t}$ is small; SGD also takes a large stepsize. However, Adam takes a small stepsize because of its large denominator $\sqrt{\mathbf{v}_t}$.

In region 3, the loss function is ``steep". Hence the gradient $\mathbf{g}_t$ and $|\mathbf{g}_t(\mathbf{x_8}) - \mathbf{g}_t(\mathbf{x_9})|$ are both very large. For this reason, an ideal optimizer should take a small stepsize. Moreover, by the design of the second order momentums in AdaBelief and Adam, they take a small stepsize in this case while SGD exploits a large stepsize.

In summary, AdaBelief fully considers all the above curvature situations, and adopts a good stepsize selection strategy in each situation. For this reason, AdaBelief has the same generalization ability as the SGD optimizer family. Though AdaBelief  retains the same regret bound guarantee $O(\sqrt{T})$ as Adam, it is interesting to explore if it can be further sped up. To this end, we propose to utilize the strong convexity and develop a new model that is able to advance the regret bound of Adabelief to logarithmic convergence in this paper.

\section{FastAdaBelief}\label{FastAdaBelief1}
In this section, we first present the detailed design of the proposed algorithm, and then analyze the theoretical guarantee of its regret bound.

\subsection{Algorithm Design}
Before presenting the proposed algorithm, we introduce some standard definitions and general assumptions, which follow previous works including \cite{Kingma2015}, \cite{Reddi2018}, \cite{Zhuang2020}, \cite{Wang2020}, \cite{Boyd2004}.

\begin{definition}\label{def1}
A function $f(\cdot):\mathcal{F}\rightarrow\mathbb{R}$ is $\sigma$-strongly convex, where $\sigma$ is a positive constant, if for all $\mathbf{x},\mathbf{y}\in\mathcal{F}$ the following equation is satisfied
\begin{equation}\label{2-1}
f(\mathbf{x}) - f(\mathbf{y}) \geq \nabla f(\mathbf{y})^{\mathsf{T}}(\mathbf{x}-\mathbf{y}) + \frac{\sigma}{2}\|\mathbf{x}-\mathbf{y}\|^2.
\end{equation}
\end{definition}

\begin{assumption}\label{ass1}
The feasible region $\mathcal{F}\in\mathbb{R}^n$ is bounded, that is, for all $\mathbf{x}, \mathbf{y}\in\mathcal{F}$,  $\max_{\mathbf{x},\mathbf{y}\in\mathcal{F}}\|\mathbf{x}-\mathbf{y}\|_{\infty}\leq D_{\infty}$, where $D_{\infty}>0$ is a constant.
\end{assumption}

\begin{assumption}\label{ass2}
For all $t\in\{1,\ldots,T\}$, the gradients of all loss functions, $\{\nabla f_t(\mathbf{x})\}_{t=1}^T$, are bounded. Specially, there exists a constant $G_{\infty}>0$ such that $\max_{\mathbf{x}\in\mathcal{F}}\|\nabla f_t(\mathbf{x})\|_{\infty} \leq G_{\infty}$.
\end{assumption}

\begin{algorithm}
\DontPrintSemicolon 
\KwIn{$\{\beta_{1t}\}_{t=1}^T, \{\beta_{2t}\}_{t=1}^T, \delta$}
\KwOut{$\mathbf{x}_{t+1}$}
\textbf{Initialize:} $\mathbf{x}_0, \mathbf{m}_0, \mathbf{s}_0$\\
\For{$t = 1 \ldots T$} {
  $t \gets t + 1$  \\
  $\alpha_t \gets \frac{\alpha}{t}$ \\
  $\mathbf{g}_t \gets \nabla f_t(\mathbf{x}_t)$ \\
  $\mathbf{m}_t \gets \beta_{1t}\mathbf{m}_{t-1} + (1-\beta_{1t})\mathbf{g}_t$ \\
  $\mathbf{s}_t \gets \beta_{2t}\mathbf{s}_{t-1} + (1-\beta_{2t})(\mathbf{g}_t - \mathbf{m}_t)^2$ \\
  $\hat{\mathbf{s}}_{t} \gets \max\{\hat{\mathbf{s}}_{t-1}, \mathbf{s}_{t}\}$ \\
  $\hat{S}_t \gets \mathrm{diag}\{\hat{\mathbf{s}}_{t}\} + \frac{\delta}{t}I_n$ \\
  $\mathbf{x}_{t+1} \gets \prod_{\mathcal{F},{\hat{S}_t}}\left( \mathbf{x}_{t} - \alpha_t \hat{S}_t^{-1}\mathbf{m}_t\right)$
}
\Return{$\mathbf{x}_{t+1}$}\;
\caption{FastAdaBelief}
\label{FastAdaBelief}
\end{algorithm}

Next we present an accelerated and accurate belief-based optimization algorithm for strongly convex functions based on the above standard definitions and assumptions, called FastAdaBelief.

The detailed design of the proposed algorithm is shown in Algorithm~\ref{FastAdaBelief}, which follows the general design of \cite{Zhuang2020}. In the proposed algorithm, $\beta_{1t}$ and $\beta_{2t}$ are time-variant non-increasing hyper-parameters, and $\delta$ is a positive constant. Moreover, the parameter of step size, $\alpha_t$, is assigned as $\alpha_t = \frac{\alpha}{t}$, where $\alpha$ is a constant. Furthermore, the gradient of loss function at time $t$, $\mathbf{g}_t$, is calculated by $\mathbf{g}_t = \nabla f_t(\mathbf{x}_t)$. Next, the proposed algorithm computes the first-order momentum, $\mathbf{m}_t$, through Exponential Moving Average (EMA) of $\mathbf{g}_t$, which is shown as follows:
\begin{equation}\label{2-2}
\mathbf{m}_t = \beta_{1t}\mathbf{m}_{t-1} + (1-\beta_{1t})\mathbf{g}_t.
\end{equation}
Then, the second-order momentum, $\mathbf{s}_t$, in the proposed algorithm is calculated by EMA of the square of the observed gradient belief $(\mathbf{g}_t-\mathbf{m}_t)$, i.e.,
\begin{equation}\label{2-3}
\mathbf{s}_t = \beta_{2t}\mathbf{s}_{t-1} + (1-\beta_{2t})(\mathbf{g}_t - \mathbf{m}_t)^2.
\end{equation}
Moreover, to satisfy the condition of convergence, i.e., $\Gamma_t = \frac{\sqrt{\mathbf{s}_t}}{\alpha_t} - \frac{\sqrt{\mathbf{s}_{t-1}}}{\alpha_{t-1}}\succeq 0$, the proposed algorithm further provides the following operation on the second-order momentum:
\begin{equation}\label{2-4}
\hat{\mathbf{s}}_{t} = \max\{\hat{\mathbf{s}}_{t-1}, \mathbf{s}_{t}\}.
\end{equation}
Furthermore, to avoid step size explosion caused by too small gradients, the proposed algorithm adds a vanishing factor $\frac{\delta}{t}$ to the second-order momentum, and obtains the following diagonal matrix:
\begin{equation}\label{2-5}
\hat{S}_t = \mathrm{diag}\{\hat{\mathbf{s}}_{t}\} + \frac{\delta}{t}I_n.
\end{equation}
Finally, the proposed algorithm updates the decision point, $\mathbf{x}_{t+1}$, conditional on the projection to the feasible region, and attains the following:
\begin{equation}\label{2-6}
\mathbf{x}_{t+1} = \prod_{\mathcal{F},{\hat{S}_t}}\left( \mathbf{x}_{t} - \alpha_t \hat{S}_t^{-1}\mathbf{m}_t\right).
\end{equation}
In general, the proposed algorithm is designed to incorporate two key enhancements compared to AdaBelief. The first  relates to the step size, which is modified to $\frac{\alpha}{t}\hat{S}^{-1}$. The motivation behind this is to satisfy the property of strongly convex optimization. Moreover, the second enhancement is to change $\beta_{2}$ of AdaBelief to $\beta_{2t}$. The time-varying parameter, $\beta_{2t}$, is set to constant $\beta_{2}$ in AdaBelief, which simplifies application and convergence proof but can lead to stepsize fluctuations. In this work, we apply $\beta_{2t}$ in its original form which  achieves good convergence~\cite{Reddi2018}.

The detailed design of our proposed algorithm has been introduced in this section. Next, we interpret why the proposed FastAdaBelief can choose  a better stepsize and leads to faster convergence. Following this, we theoretically prove that when strong convexity of loss functions holds,  our proposed algorithm has a guaranteed regret bound, which is much better than AdaBelief.

\subsection{Why FastAdaBelief can choose a better stepsize?}
From the design o $\hat{S}_t$ in FastAdaBelief, our algorithm adds a vanishing factor to the stepsize, which was originally considered to meet the strongly convex condition, but unexpectedly brings significant benefits to the choice of the step size. If we let $\Delta$ denote the stepsize, then stepsizes of SGD, Adam, SAdam, AdaBelief, and FastAdaBelief are shown in the following:

\begin{align}
   &\Delta_t(SGD) = \alpha\mathbf{m}_t; \nonumber \\
   &\Delta_t(Adam) = \alpha\mathbf{m}_t/\big(\sqrt{t\mathbf{v}_t}\big); \nonumber \\
   &\Delta_t(SAdam) = \alpha\mathbf{m}_t/\big(t\mathbf{v}_t+\delta\big); \nonumber \\
   &\Delta_t(AdaBelief) = \alpha\mathbf{m}_t/\big(\sqrt{t\mathbf{s}_t}\big); \nonumber \\
   &\Delta_t(FastAdaBelief) = \alpha\mathbf{m}_t/\big(t\mathbf{s}_t+\delta\big). \nonumber
\end{align}
It can also be seen from Figure~\ref{fig0}, that in regions 1 and 2, although the step size of FastAdaBelief is slightly smaller than that of AdaBelief, it is still consistent with the optimal choice. FastAdaBelief has large step sizes in both region 1 and 2, but SAdam takes small step sizes. Thus FastAdaBelief outperforms SAdam with respect to generalization ability. Importantly, the step size of FastAdaBelief decays in general on the order of $O(1/t)$ that allows the optimal solution to be approximated at a smaller step size later in the training process without unnecessary oscillations. In addition, in region 3, the ideal optimizer would prefer  a small stepsize. FastAdaBelief takes a smaller  stepsize than AdaBelief in this region, which is due to  the addition of vanishing factors.

In summary, a comparison of stepsize selection by FastAdaBelief, AdaBelief, SAdam, Adam and SGD can be seen in Table \ref{tableStepsize}. This analysis shows that FastAdaBelief, like AdaBelief, is in line with the choice of the ideal optimizer and therefore can lead to better performance than other mainstream optimizers. Moreover, FastAdaBelief has a smaller step size than that of AdaBelief in the later stage of training, which allows the optimizer to approximate the optimal solution more steadily. Therefore, FastAdaBelief, like AdaBelief, has a better generalization ability than other algorithms when training deep models.

\begin{table*}[t]
  \caption{Stepsize selected by the ideal optimizer, FastAdaBelief, AdaBelief, SAdam, Adam and SGD in three regions of Figure~\ref{fig0}, where $L$ denotes a large value and $S$ denotes a small value. FastAdaBelief is more in line with the ideal optimizer.}
  \label{tableStepsize}
  \centering
  \begin{tabular}{lccccc}
    \toprule
    \textbf{Stepsize}      &\textbf{Region 1}     &\textbf{Region 2}   &\textbf{Region 3} &\textbf{Later Period}\\
    \midrule
    $\Delta_t(ideal)$      &$L$                   &$L$                 &$S$  &steady\\
    \midrule
    $\Delta_t(SGD)$        &$S$                   &$L$                 &$L$  &oscillating \\
    \midrule
    $\Delta_t(Adam)$       &$L$                   &$S$                 &$S$  &oscillating \\
    \midrule
    $\Delta_t(SAdam)$      &$L$                   &$S$                 &$S$  &steady \\
    \midrule
    $\Delta_t(AdaBelief)$   &$L$                  &$L$                 &$S$  &oscillating \\
    \midrule
    $\Delta_t(FastAdaBelief)$  &$L$               &$L$                 &$S$  &steady\\
    \bottomrule
  \end{tabular}
\end{table*}

\subsection{Theoretical Guarantee}
In this section, we first review some convergence conditions as developed by Reddi \cite{Reddi2018}, which solves the convergence issue for Adam \cite{Kingma2015}. Let $\{\beta_{2t}\}$ satisfy the following conditions:

\paragraph{Condition 1} For some $\zeta>0$ and all $t\in\{1,\ldots,T\}$, $j\in\{1,\ldots,n\}$, we have that
\begin{equation*}
\frac{\sqrt{t}}{\alpha}\sqrt{ \sum_{j=1}^t\prod_{k=1}^{t-j}\beta_{2(t-k+1)}(1-\beta_{2j})g_{j,i}^2 } \geq \frac{1}{\zeta}\sqrt{\sum_{j=1}^t g_{j,i}^2}.
\end{equation*}

\paragraph{Condition 2.} For all $t\in\{1,\ldots,T\}$ and $i\in\{1,\ldots,n\}$, we have that
\begin{equation*}
\frac{\sqrt{t}}{\alpha} s_{t,i}^{1/2} \geq \frac{\sqrt{t-1}}{\alpha} s_{t-1,i}^{1/2}.
\end{equation*}

As a matter of fact, Condition 1 is an important and standard condition for convergence analysis of adaptive momentum algorithms, such as Adam and AdaBelief. Furthermore, the intrinsic motivation for Condition 2 is to follow the key condition of SGD, where its step size $\frac{\alpha}{\sqrt{t}}$ satisfies that $\frac{\sqrt{t}}{\alpha} - \frac{\sqrt{t-1}}{\alpha} \geq 0, \forall t\in [T]$. For this reason, we also follow this motivation, and propose the following conditions with minor modifications:

\paragraph{Condition 3.} For some $\zeta>0$ and all $t\in\{1,\ldots,T\}$, $j\in\{1,\ldots,n\}$, we have that
\begin{equation}\label{c-1}
t\sum_{j=1}^t\prod_{k=1}^{t-j}\beta_{2(t-k+1)}(1-\beta_{2j})g_{j,i}^2 \geq \frac{1}{\zeta}\sum_{j=1}^t g_{j,i}^2.
\end{equation}

\paragraph{Condition 4.} For all $t\in\{1,\ldots,T\}$ and $i\in\{1,\ldots,n\}$, we have that
\begin{equation}\label{c-2}
0\leq\frac{t}{\alpha} s_{t,i}^{1/2} - \frac{t-1}{\alpha} s_{t-1,i}^{1/2}\leq\sigma(1-\beta_1).
\end{equation}

The details on Condition 3 and Condition 4 are shown in Appendix \ref{appendixb}.

Now, we present the main results in the following for the convergence analysis when Conditions 3 and 4 are satisfied.

\begin{theorem}\label{thm1}
Suppose that Assumptions \ref{ass1} and \ref{ass2} are satisfied, Conditions 3 and 4 hold, and loss functions $f_t(\cdot)$ are $\sigma$-strongly convex. Moreover, let parameter sequences $\{\beta_{1t}\}, \{\beta_{2t}\}$ and $\{\alpha_{t}\}$ are generated by the proposed algorithm, where $\beta_{1t}=\beta_1\lambda^t, \beta_1\in[0,1), \lambda_1\in[0,1), \beta_{2t}\in[0,1), \delta>0, t\in\{1,\ldots,T\}$. For decision point $\mathbf{x}_t$ generated by the proposed algorithm, we have the following upper bound of the regret
\begin{align*}
R(T) \leq &\frac{n\delta D_{\infty}^2}{2\alpha(1-\beta_1)}+\frac{D_{\infty}^2(G_{\infty}+\delta)}{2\alpha}\sum_{i=1}^n\sum_{t=1}^T\frac{\beta_{1t}}{1-\beta_{1t}}t  \nonumber \\
&+ \frac{\alpha\zeta}{\varpi^2(1-\beta_1)^3}\sum_{i=1}^{n}\log\left(\frac{1}{\zeta\delta}\|g_{1:T,i}\|^2+1\right).
\end{align*}
\end{theorem}

The proof of Theorem \ref{thm1} is provided in Appendix \ref{app1}. Accordingly, Theorem \ref{thm1} implies that our proposed algorithm converges with $O\big(\sum_{i=1}^n\log\big(\left\|g_{1:T,i}\right\|^2\big)\big)$ regret bound in the case of  strong convexity. Moreover, the regret bound of the worst case is $O(n\log T)$. In addition, the bound of the regret can be more tighter if the gradients are sparse or small such that $\left\|g_{1:T,i}\right\|^2\ll TG_{\infty}^2 $.

\begin{corollary}\label{cor1}
Letting $\beta_{1t}=\beta_1\lambda^t$, where $\lambda\in(0,1)$ in Theorem \ref{thm1},  we have the following upper bound of the regret
\begin{align*}
R(T) &\leq \frac{n\delta D_{\infty}^2}{2\alpha(1-\beta_1)} +\frac{n\beta_{1}\lambda D_{\infty}^2(G_{\infty}+\delta)}{2\alpha(1-\beta_{1})(1-\lambda)^2} \nonumber \\
& + \frac{\alpha\zeta}{\varpi^2(1-\beta_1)^3}\sum_{i=1}^{n}\log\left(\frac{1}{\zeta\delta}\|g_{1:T,i}\|^2+1\right).
\end{align*}
\end{corollary}

The above Corollary \ref{cor1} also implies that our proposed algorithm has a convergence guarantee $O(n\log T)$ for condition $\beta_{1t}=\beta_{1}\lambda^t, \lambda\in(0,1), t\in\{1,\ldots,T\}$. Then, our proposed algorithm executes with $\lim_{T\rightarrow +\infty}\frac{R(T)}{T} = 0$. Therefore, our proposed algorithm converges when loss functions are strongly convex, and its theoretical proof is provided in Appendix \ref{app1}. In order to verify the performance of our algorithm in specific applications, we present a series of experiments on benchmark public datasets in the next section.
\section{Experiments}
\label{Experiments}
In this section, we conduct two groups of experiments to verify that our proposed algorithm works excellently for benchmark optimization problems in  cases of strong convexity and non-convexity. In the first group, we consider a strongly convex optimization problem of mini-batch $\ell_2$-regularized softmax regression; in the second group, we apply our algorithm to non-convex cases of deep training tasks with the traditional softmax function. To be specific, the traditional softmax function is generally convex but not strongly convex when the data is uniformly distributed.  For this reason, the softmax function  does not often satisfy convexity in deep neural network applications due to the sparsity of data and the nonlinearity of deep neural networks \cite{Song2020,Luox2019}. Therefore, deep learning tasks are generally non-convex optimization problems \cite{Liu2016}. To examine the effectiveness of FastAdaBelief in real scenarios, we intentionally conduct a second group of experiments on image classification with CNN and  language modeling with LSTM respectively, both of which are commonly seen in practice. In all of our experiments, the source codes are implemented in the torch 1.1.0 module of python 3.6 and executed on $4\times1080$ti GPUs. Furthermore, we compare FastAdaBelief with the other algorithms in both experiments, including SGD \cite{Sutskever2013}, Adam \cite{Kingma2015}, Yogi \cite{Zaheer2018}, AdaBound \cite{Luo2019}, AdaBelief \cite{Zhuang2020} and SAdam \cite{Wang2020}. We independently  execute the experiments  5 times, and finally report the top-1 of them, which follows \cite{Zhuang2020},\cite{Zaheer2018}.

\subsection{Hyperparameter Tuning}
We perform the following hyperparameter tuning in experiments of image classification and language modeling. To be fair, we initialize the decision variables and momentum of each algorithm to $\mathbf{x}_0 = \mathbf{0}, \mathbf{m}_0 = \mathbf{0}, \mathbf{v}_0 = \mathbf{0}$, and $\mathbf{s}_0 = \mathbf{0}$. Moreover, we choose the parameters of each algorithm exactly as suggested in the original papers. Specifically, the parameter settings for each algorithm are as follows.

\emph{SGD:} We follow the standard settings of ResNet \cite{He2016} and DenseNet \cite{Liu2017}, and set the momentum as $0.9$. We choose the learning rate from $\{10.0, 1.0, 0.1, 0.01, 0.001\}$.

\emph{Adam:} We adopt the same parameter setting as the original article \cite{Kingma2015} where the first-order momentum $\beta_1$ is set to $0.9$, and the second-order momentum $\beta_2$ is set to $0.999$. Moreover, the step size $\alpha_t$ is set to $\alpha / \sqrt{t}$, where $\alpha$ is chosen from $\{0.1, 0.01, 0.001, 0.0001\}$.

\emph{Yogi:} Following \cite{Zaheer2018}, we set the first-order momentum $\beta_1$ to $0.9$ and set the second-order momentum $\beta_2$ to $0.999$. In addition, the step size $\alpha_t$ is set to $\alpha / \sqrt{t}$, where $\alpha$ is chosen from $\{0.1, 0.01, 0.001, 0.0001\}$.

\emph{AdaBound:} We directly apply the default hyper-parameters following \cite{Luo2019} for AdaBound (i.e., $\beta_1 = 0.9$ and $\beta_2 = 0.999$). Moreover, the step size $\alpha_t$ is set to $\alpha / \sqrt{t}$, where $\alpha$ is chosen from $\{0.1, 0.01, 0.001, 0.0001\}$.

\emph{AdaBelief:} We use the default hyperparameters as suggested in \cite{Zhuang2020}, i.e., $\beta_1 = 0.9$, $\beta_2 = 0.999$, and $\epsilon=10^{-8}$. In addition, the step size $\alpha_t$ is set to $\alpha / \sqrt{t}$, where $\alpha$ is chosen from $\{0.1, 0.01, 0.001, 0.0001\}$.

\emph{SAdam:} We set the hyperparameters  by following \cite{Wang2020}, i.e., $\beta_1 = 0.9$, $\beta_{2t} = 1- \frac{0.9}{t}$. The step size $\alpha_t$ is set to $\alpha / t$, where $\alpha$ is chosen from $\{0.1, 0.01, 0.001, 0.0001\}$.

\emph{FastAdaBelief:} We adopt the same hyperparameters as SAdam: $\beta_1 = 0.9$, $\beta_{2t} = 1- \frac{0.9}{t}$. Moreover, the step size is set to $\alpha / t$, where $\alpha$ is chosen from $\{0.1, 0.01, 0.001, 0.0001\}$.

It can be seen that, for fair comparison, all the above algorithms basically follow a similar parameter setting and  maintain the parameter suggestions provided in the original algorithms.

\subsection{Datasets}
In the experiments of CNN based image classification,  we perform evaluations on the benchmark CIFAR-10 dataset. Moreover, we apply the algorithms on three standard baseline models, i.e., DenseNet-121, ResNet-34, and VGG-11. DenseNet-121 is a dense convolutional network, which connects each layer to all other layers feed-forwardly; ResNet-34 is a residual learning framework; VGG-11 is a deep network using an architecture with small convolution filters. In the LSTM based language modeling experiments, we test the various algorithms on Penn Treebank dataset. Furthermore, we compare the algorithms in 1,2,3-layer LSTM models. For clarity, we show the summary of datasets and architectures used in our experiments in Table~\ref{table2}.
\begin{table*}
  \caption{Datasets and architectures used in our experiments.}
  \label{table2}
  \centering
  \begin{tabular}{llll}
    \toprule
    Task & Dataset  & Architecture \\
    \midrule
    Image Classification & SVHN       & $\ell_2$-regularized softmax regression     \\
    Image Classification & CIFAR-10   & 4-layers CNN,DenseNet-121, ResNet-34, VGG-11 \\
    Image Classification & CIFAR-100  & 4-layers CNN      \\
    Language Modeling & Penn Treebank & 1,2,3-Layer LSTM. \\
    \bottomrule
  \end{tabular}
\end{table*}

\subsection{Optimization with  Strong Convexity}

\begin{figure*}
    \centering
		\includegraphics[scale=.37]{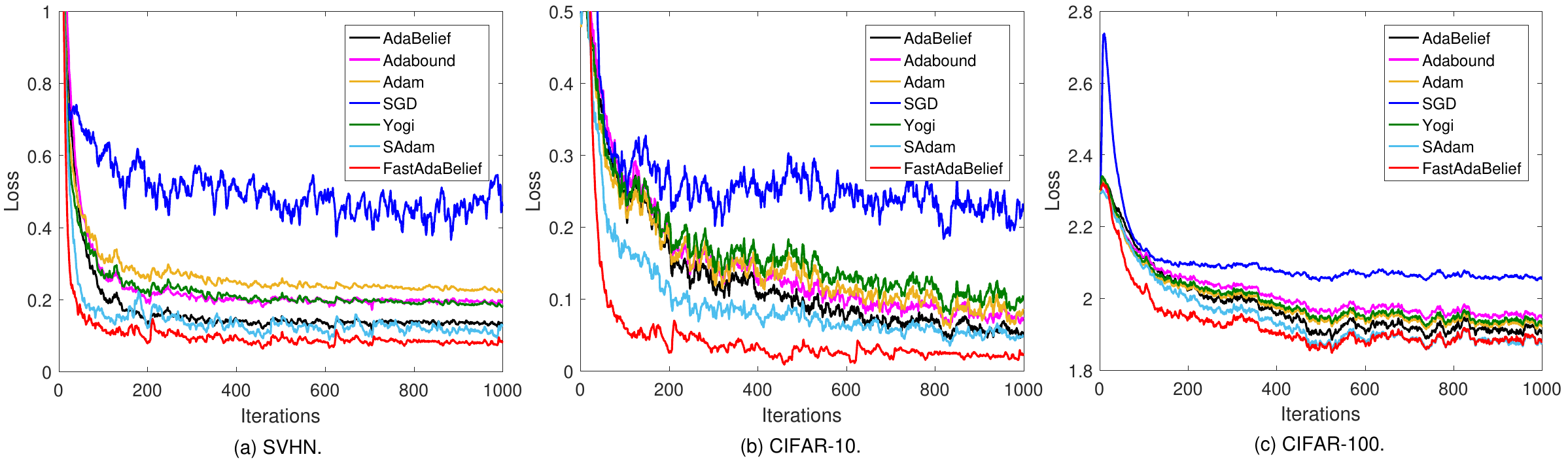}
	    \caption{Loss \emph{v.s.} iterations for mini-batch $\ell_2$-regularized softmax regression (\textbf{strongly convex}). FastAdaBelief converges the quickest.}
	    \label{fig1}
\end{figure*}

\begin{figure*}
    \centering
		\includegraphics[scale=.31]{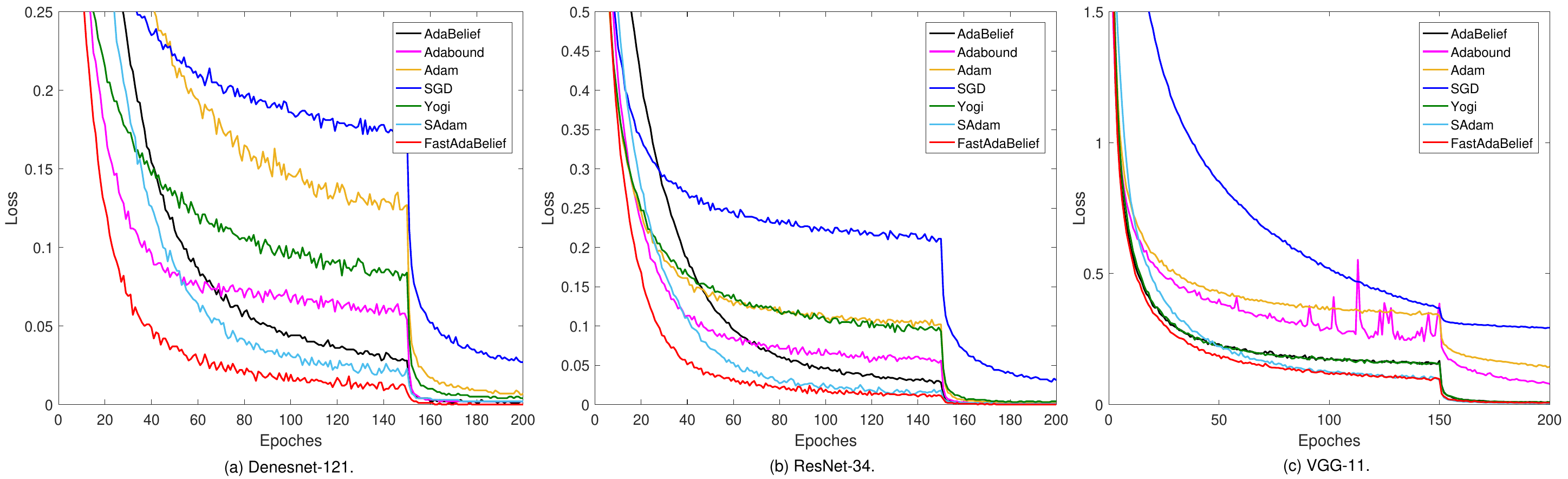}
	    \caption{Comparison of loss of SGD, Adam, AdaBound, Yogi, AdaBelief, SAdam and FastAdaBelief on CIFAR-10. FastAdaBelief converges the quickest.}
	    \label{fig2}
\end{figure*}

\begin{figure*}
    \centering
		\includegraphics[scale=.38]{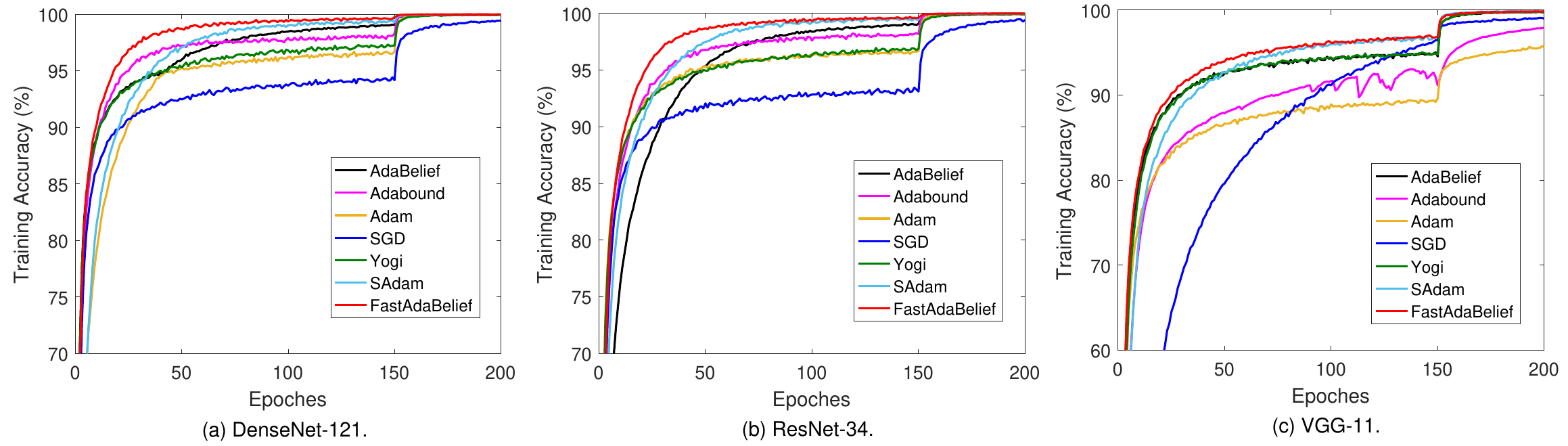}
	    \caption{Comparison of training accuracy of SGD, Adam, AdaBound, Yogi, AdaBelief, SAdam and FastAdaBelief on CIFAR-10. FastAdaBelief achieves the highest training accuracy.}
	    \label{fig3}
\end{figure*}

\begin{figure*}
    \centering
		\includegraphics[scale=.4]{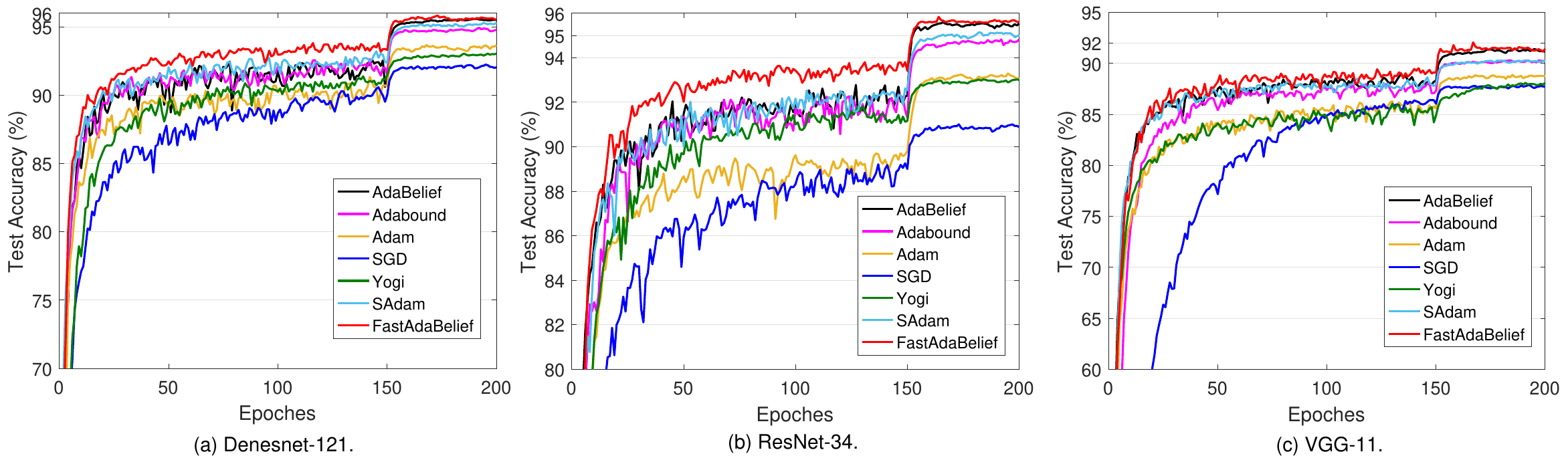}
	    \caption{Comparison of test accuracy of SGD, Adam, AdaBound, Yogi, AdaBelief, SAdam and FastAdaBelief on CIFAR-10. FastAdaBelief achieves the highest test accuracy.}
	    \label{fig4}
\end{figure*}

\begin{figure*}
    \centering
		\includegraphics[scale=.32]{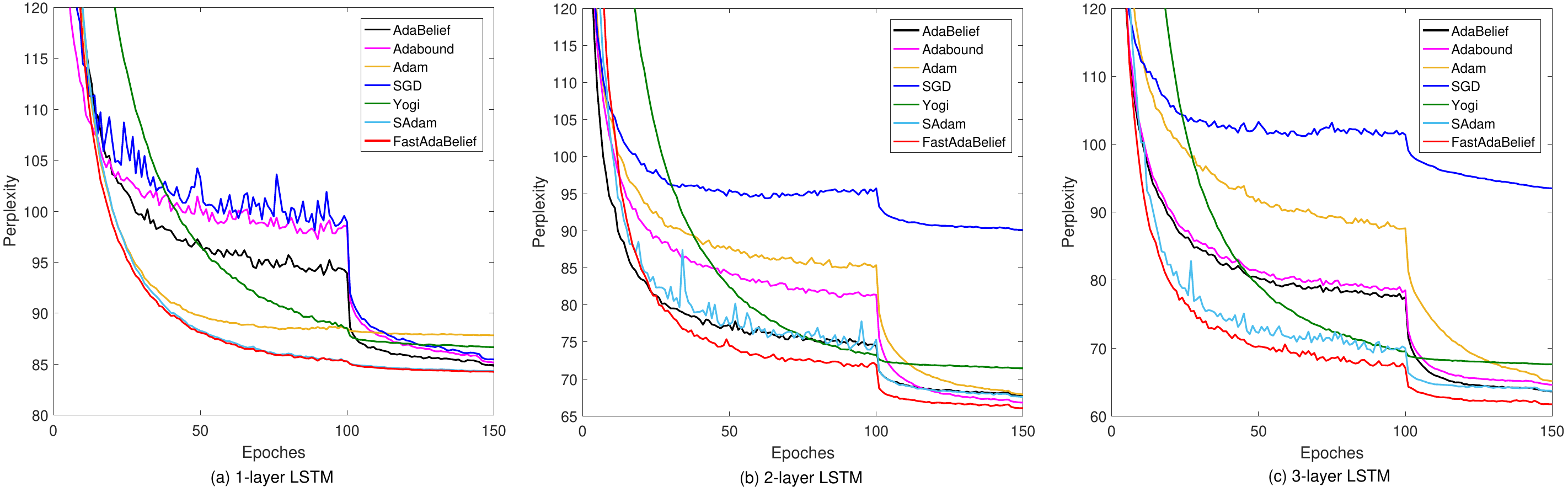}
	    \caption{Comparison of perplexity of SGD, Adam, AdaBound, Yogi, AdaBelief, SAdam and FastAdaBelief (\textbf{Lower} is better)  on Penn Treebank. FastAdaBelief converges the quickest.}
	    \label{fig5}
\end{figure*}

In this group of experiments, we consider a mini-batch task. In round $t$ of this task, the optimizer receives a mini-batch of training samples denoted by $\{\mathbf{x}_m,y_m\}_{i=1}^m$, where $m$ is the batch size, $K$ is the number of classes,  and $y_i\in[K]$ and $\forall i\in[m]$. Then the optimizer generates decision vectors denoted by $\{\mathbf{w}_i,b_i\}_{i=1}^K$. Finally, the generated result suffers a loss. The loss function is then given as \begin{align}
J(\mathbf{w}) = &-\frac{1}{m}\sum_{i=1}^m \log\left(\frac{e^{\mathbf{w}_{y_i}^{\top} \mathbf{x}_i + b_{y_i}}}{\sum_{j=1}^{K} e^{\mathbf{w}_j^{\top} \mathbf{x}_i + b_j}}\right) + \sigma_1\sum_{k=1}^K\|\mathbf{w}_k\|^2\nonumber \\
& + \sigma_2\sum_{k=1}^K b_k^2.
\end{align}
In our experiments, we set parameters $\sigma_1$ and $\sigma_2$ both to $0.01$. In addition, we conduct the experiment on loss \emph{v.s.} iterations.  The results of this experiment are shown in Figure~\ref{fig1}. As clearly observed, the loss of our proposed  algorithm  decreases the quickest and  FastAdaBelief leads to the best convergence in all mainstream  algorithms. As guaranteed theoretically, the strongly convex optimization algorithms (such as FastAdaBelief and SAdam) outperform convex optimization algorithms (like Adam, AdaBelief, etc.) in the strongly convex case. When we inspect the difference between the two  strongly convex optimization algorithms,  FastAdaBelief generates much lower losses (particularly on CIFAR10 and CIFAR100) than SAdam, which echos the advantages of FastAdaBelief over SAdam in relation to the generalization ability.

\begin{table*}
  \caption{Test perplexity (\textbf{lower} is better) of 1,2,3-layer LSTM on Penn Treebank. Note that the two best performing algorithms are marked in bold.}
  \label{table3}
  \centering
  \begin{tabular}{llllllll}
    \toprule
    Model       & SGD   & Adam & AdaBound & Yogi  & AdaBelief & SAdam & FastAdaBelief    \\
    \midrule
    1-layer LSTM& 85.07 & 84.28& 84.78    & 86.59 & 84.21     & \textbf{84.19} & \textbf{84.18} \\
    \midrule
    2-layer LSTM& 67.42 & 67.27& 67.53    & 71.33 & \textbf{66.29}     & 68.11 & \textbf{66.08} \\
    \midrule
    3-layer LSTM& 63.58 & 64.28& 63.58    & 67.51 & \textbf{61.23}     & 64.71 & \textbf{61.21} \\
    \bottomrule
  \end{tabular}
\end{table*}

\subsection{Training DNN with Non-convexity}
FastAdaBelief enjoys theoretical superiority over other mainstream optimizers when strong convexity holds. On the empirical side, the current DNNs may however adopt loss functions that are typically not strongly convex (e.g. only convex). Thus it is both interesting and important to investigate if the proposed fast algorithm can still work well. For this purpose, we next conduct a series of  empirical studies on tasks of image classification and language modeling.

\subsubsection{Image Classification}
In the experiments of image classification, we take CIFAR-10 as one typical example and compare the various algorithms with DenseNet-121, ResNet-34 and VGG-11. First, we compare the convergence rate for all the algorithms used in our experiment. Such results are reported in Figure~\ref{fig2}.  As clearly observed, though the strong convexity may not hold, FastAdaBelief still leads to  remarkable convergence, which is consistently faster than all the other algorithms. In comparison, SAdam also converges well, which empirically demonstrates the power of  strongly convex algorithms. Furthermore, SGD converges the slowest; Adam and Adabelief are also much slower than both SAdam and FastAdaBelief. All these empirical results are consistent with the theoretical analysis as discussed earlier in Section~\ref{related} and Section~\ref{FastAdaBelief1} though the loss functions are not strongly convex.


Second, we record the training and test accuracy curves of all algorithms executed in our experiments, which are shown in Figure~\ref{fig3} and Figure~\ref{fig4}. We can see that FastAdaBelief outperforms other comparison algorithms in 200 epochs on DenseNet-121, ResNet-34 and VGG-11. Specifically, FastAdaBelief demonstrates much faster convergence as well as the highest accuracy within 200 epochs compared to all other algorithms on the three baseline DNN models. Additionally, it is evident that AdaBelief and FastAdaBelief generally lead to best accuracy in the 200 epoch, which  verifies the excellent generalization ability of belief-based adaptive algorithms. Note that SGD did not converge due to its slow convergence rate though it was able to catch up with the accuracy of AdaBelief and FastAdaBelief in the long run.



To sum up, the experiments of image classification with DenseNet-121, ResNet-34 and VGG-11 on CIFAR-10 validate the fast convergence rate and excellent accuracy performance of FastAdaBelief even when strong convexity does not hold in the loss functions.


\subsubsection{Language Modeling}
We also conduct a group of experiments on the language modeling task. In this group of experiments, we use a classic recurrent network (i.e., LSTM) and an open dataset (i.e., Penn Treebank). In line with previous works~\cite{Zhuang2020}, \cite{Wang2020}, \cite{Shuang2020}, \cite{Huang2010}, we take the perplexity to measure the performance of all algorithms under comparison. Note that a lower perplexity is better.

The perplexity curves of all the algorithms are shown in Figure~\ref{fig5}. In the figure, once again we can see that FastAdaBelief performs similar to SAdam on 1-layer LSTM, and these two algorithms both perform better than the rest of the algorithms. However, on the 2,3-layer LSTM, FastAdaBelief performs better than SAdam and the other algorithms. Moreover, the  perplexity of FastAdaBelief decreases the fastest among all the algorithms, which further  validates the convergence analysis of FastAdaBelief.

We also summarize the test perplexities of all algorithms compared in this group of experiments, which are shown in Table~\ref{table3}. From the table, we can see that FastAdaBelief attains superior performance to the other algorithms on all three models (i.e., 1,2,3-layer LSTM). In summary, FastAdaBelief retains both excellent generalization ability and fast convergence rate for language modeling tasks even when loss functions are not strongly convex. These empirical results are very encouraging, suggesting that FastAdaBelief has high potential  to be widely applied in real scenarios.

\subsection{Discussion on Convexity}
In the above, we focus on the convergence benefits of algorithms under strongly convex conditions and propose a strongly convex optimization algorithm. We prove that it converges faster than convex algorithms, such as Adam and AdaBelief, in strongly convex cases, which is also verified in the above experimental part. Moreover, the proposed algorithm empirically exhibits much faster convergence than Adam and AdaBelief in non-convex cases as well.

In the case of convexity, we provide the regret bound proof of FastAdaBelief to fully understand  its advantages and disadvantages. The proof presented in Appendix \ref{app2} shows that FastAdaBelief actually converges slower than certain special  convex algorithms, such as Adabelief and Adam, in convex cases. Therefore,  we suggest to use the proposed FastAdaBelief in strongly convex and non-convex cases while convex algorithms e.g. Adabelief can be used in convex cases.

\section{Conclusion and Future Work}
In this paper, we made a first attempt and presented an affirmative answer to the open question of whether AdaBelief can be further improved with respect to its convergence rate under the strongly convex condition. Specifically, we exploited strong convexity and proposed a novel algorithm named FastAdaBelief,  which  exhibits an even faster data-dependent regret bound of  $O(\log T)$ while maintaining excellent generalization ability. In light of our theoretical findings, we carried out a series of empirical studies which validated the superiority of our proposed algorithm. Importantly, we showed that FastAdaBelief converged the fastest in both strong convexity and non-convexity cases, hence demonstrating its significant potential as a new benchmark model that can be widely utilised in various scenarios.

In our current work, we exploited the strong convexity of FastAdaBelief, and empirically demonstrated its excellent generalization as well as fast convergence even when the loss functions are non-convex. We believe this may be partially attributable to the vanishing factor $\delta/t$, as also engaged in the second order moment that enables a closer approximation to an ideal step size. However, it remains unclear why this may happen strictly in theory. We will explore this in future work.  In addition, whilst research on the sparsity of samples can improve the convergence rate of SGD, as demonstrated in \cite{Luo1,Luo2,Luo3}, it remains unclear whether sparse samples will further improve FastAdaBelief's convergence rate. We also leave this investigation as future work.


\section*{Acknowledgment}
This work was supported by the Hundred Talents Program of Chinese Academy of Sciences under grant No. Y9BEJ11001, and also supported by the innovation workstation of Suzhou Institute of Nano-Tech and Nano-Bionics (SINANO) under grant No. E010210101. Huang would like to acknowledge the support of National Natural Science Foundation of China under no.61876155, and Jiangsu Science and Technology Programme under no. BE2020006-4. Hussain would like to acknowledge the support of the UK Engineering and Physical Sciences Research Council (EPSRC) - Grants Ref. EP/M026981/1, EP/T021063/1, EP/T024917/1.




%

\appendices
\section{Convergence Analysis in Strongly Convex Online Optimization}\label{app1}
Before presenting the proof of Theorem \ref{thm1}, we first review the following Lemma \ref{lem1}.
\begin{lemma}\label{lem1}
\cite{McMahan2010} For all $M\in\mathcal{M}_{+}^n$ and convex feasible region $\mathcal{F}\in\mathbb{R}^{n}$, let $$\mathbf{y}_1 = \min_{\mathbf{x}\in\mathcal{F}}\left\|M^{1/2}(\mathbf{x}-\mathbf{z}_1)\right\|,$$ and $$\mathbf{y}_2 = \min_{\mathbf{x}\in\mathcal{F}}\left\|M^{1/2}(\mathbf{x}-\mathbf{z}_2)\right\|,$$ then we obtain the following:
$$\left\|M^{1/2}(\mathbf{y}_1 - \mathbf{y}_2)\right\| \leq \left\|M^{1/2}(\mathbf{z}_1 - \mathbf{z}_2)\right\|.$$
\end{lemma}

\textbf{Theorem 1}\textit{
Suppose that Assumptions \ref{ass1} and \ref{ass2} are satisfied, Conditions 3 and 4 hold, and loss functions $f_t(\cdot)$ are $\sigma$-strongly convex. Moreover, let parameter sequences $\{\beta_{1t}\}, \{\beta_{2t}\}$ and $\{\alpha_{t}\}$ are generated by the proposed algorithm, where $\beta_{1t}=\beta_1\lambda^t, \beta_1\in[0,1), \lambda_1\in[0,1), \beta_{2t}\in[0,1), \delta>0, t\in\{1,\ldots,T\}$. For decision point $\mathbf{x}_t$ generated by the proposed algorithm, we have the following upper bound of the regret
\begin{footnotesize}
\begin{align*}
R(T) \leq &\frac{n\delta D_{\infty}^2}{2\alpha(1-\beta_1)}+\frac{D_{\infty}^2(G_{\infty}+\delta)}{2\alpha}\sum_{i=1}^n\sum_{t=1}^T\frac{\beta_{1t}}{1-\beta_{1t}}t  \nonumber \\
&+ \frac{\alpha\zeta}{\varpi^2(1-\beta_1)^3}\sum_{i=1}^{n}\log\left(\frac{1}{\zeta\delta}\|g_{1:T,i}\|^2+1\right).
\end{align*}
\end{footnotesize}}

\paragraph{\emph{Proof}} By the updating method of decision variable, i.e., Equation \eqref{2-6}, we have:
\begin{footnotesize}
\begin{align}\label{a-1}
\mathbf{x}_{t+1} = \prod_{\mathcal{F},{\hat{S}_t}}\left(\mathbf{x}_{t} - \alpha_t \hat{S}_t^{-1}\mathbf{m}_t\right) = \min_{\mathbf{x}\in\mathcal{F}}\left\|\hat{S}_t^{1/2}\left[\mathbf{x}-(\mathbf{x}_t - \alpha_t \hat{S}_t^{-1}\mathbf{m}_t)\right]\right\|.
\end{align}
\end{footnotesize}
From the definitions of $\mathbf{x}^*$ and projection $\prod (\cdot)$, we have that $\mathbf{x}^*=\prod_{\mathcal{F},\hat{S}_t}(\mathbf{x}^*)=\min_{\mathbf{x}\in\mathcal{F}}(\mathbf{x}-\mathbf{x}^*)$. In addition, if we apply Lemma \ref{lem1}, let $\mathbf{y}_1 = \mathbf{x}_{t+1}, \mathbf{y}_2 = \mathbf{x}^*$,  by the update rules of $\mathbf{x}_{t}$ and $\mathbf{m}_t$, we obtain the following:
\begin{footnotesize}
\begin{align}\label{a-2}
 &\left\|\hat{S}_t^{1/2}\left(\mathbf{x}_{t+1}-\mathbf{x}^*\right)\right\|^2 \leq \left\|\hat{S}_t^{1/2}\left(\mathbf{x}_t-\alpha_t \hat{S}_t^{-1}\mathbf{m}_t -\mathbf{x}^* \right)\right\|^2 \nonumber \\
 &= \left\|\hat{S}_t^{1/2}\left(\mathbf{x}_t -\mathbf{x}^* \right)\right\|^2 - \hat{S}_t\left\langle\mathbf{x}_t -\mathbf{x}^*, \alpha_t \hat{S}_t^{-1}\mathbf{m}_t\right\rangle + \left\|\alpha_t \hat{S}_t^{-1/2}\mathbf{m}_t\right\|^2 \nonumber \\
 &= \left\|\hat{S}_t^{1/2}\left(\mathbf{x}_t -\mathbf{x}^* \right)\right\|^2 + \alpha_t^2\left\|\hat{S}_t^{-1/2}\mathbf{m}_t\right\|^2 - 2\alpha_t\left\langle\mathbf{x}_t -\mathbf{x}^*, \mathbf{m}_t\right\rangle \nonumber \\
 &= \left\|\hat{S}_t^{1/2}\left(\mathbf{x}_t -\mathbf{x}^* \right)\right\|^2 + \alpha_t^2\left\|\hat{S}_t^{-1/2}\mathbf{m}_t\right\|^2 \nonumber \\
 &- 2\alpha_t\left\langle\mathbf{x}_t -\mathbf{x}^*, \beta_{1t}\mathbf{m}_{t-1} + (1-\beta_{1t})\mathbf{g}_t\right\rangle.
\end{align}
\end{footnotesize}

Next, rearranging equation \eqref{a-2}, we have that
\begin{footnotesize}
\begin{align}\label{a-3}
&\left\langle\mathbf{g}_t, \mathbf{x}_t -\mathbf{x}^*\right\rangle \leq \frac{\left[\left\|\hat{S}_t^{1/2}\left(\mathbf{x}_t -\mathbf{x}^* \right)\right\|^2 - \left\|\hat{S}_t^{1/2}\left(\mathbf{x}_{t+1}-\mathbf{x}^*\right)\right\|^2\right]}{2\alpha_t(1-\beta_{1t})} \nonumber \\
&+ \frac{\alpha_t}{2(1-\beta_{1t})}\left\|\hat{S}_t^{-1/2}\mathbf{m}_t\right\|^2 \underbrace{- \frac{\beta_{1t}}{1-\beta_{1t}}\left\langle\mathbf{m}_{t-1}, \mathbf{x}_t -\mathbf{x}^*\right\rangle}_{(a)}.
\end{align}
\end{footnotesize}
Applying Young's inequality (i.e., $\langle a, b\rangle \leq \frac{a^2 \epsilon}{2} + \frac{b^2}{2\epsilon}, \forall\epsilon>0$) into the term $(a)$ of Equation \eqref{a-3}, and considering $\alpha_t >0, \beta_{1t}\in[0,1)$, we can attain
\begin{footnotesize}
\begin{align}\label{a-4}
&(a) = -\frac{\beta_{1t}}{1-\beta_{1t}}\left\langle\mathbf{m}_{t-1}, \mathbf{x}_t -\mathbf{x}^*\right\rangle \leq \frac{\alpha_t\beta_{1t}}{2(1-\beta_{1t})}\left(\hat{S}_t^{-1/2}\mathbf{m}_{t-1}\right)^2 \nonumber \\
&+ \frac{\beta_{1t}}{2\alpha_t(1-\beta_{1t})}\left(\hat{S}_t^{1/2}(\mathbf{x}_t -\mathbf{x}^*)\right)^2.
\end{align}
\end{footnotesize}
Furthermore, applying Cauchy-Schwartz inequality into Equation \eqref{a-4}, we have
\begin{footnotesize}
\begin{align}\label{a-5}
(a) \leq \frac{\alpha_t\beta_{1t}}{2(1-\beta_{1t})}\left\|\hat{S}_t^{-1/2}\mathbf{m}_{t-1}\right\|^2 + \frac{\beta_{1t}}{2\alpha_t(1-\beta_{1t})}\left\|\hat{S}_t^{1/2}(\mathbf{x}_t -\mathbf{x}^*)\right\|^2.
\end{align}
\end{footnotesize}
Then, plugging Equation \eqref{a-5} into Equation \eqref{a-3}, we obtain the following
\begin{footnotesize}
\begin{align}\label{a-6}
&\left\langle\mathbf{g}_t, \mathbf{x}_t -\mathbf{x}^*\right\rangle \leq \frac{\left[\left\|\hat{S}_t^{1/2}\left(\mathbf{x}_t -\mathbf{x}^* \right)\right\|^2 - \left\|\hat{S}_t^{1/2}\left(\mathbf{x}_{t+1}-\mathbf{x}^*\right)\right\|^2\right]}{2\alpha_t(1-\beta_{1t})} \nonumber \\
&+ \frac{\alpha_t}{2(1-\beta_{1t})}\left\|\hat{S}_t^{-1/2}\mathbf{m}_t\right\|^2 + \frac{\alpha_t\beta_{1t}}{2(1-\beta_{1t})}\left\|\hat{S}_t^{-1/2}\mathbf{m}_{t-1}\right\|^2 \nonumber \\
&+ \frac{\beta_{1t}}{2\alpha_t(1-\beta_{1t})}\left\|\hat{S}_t^{1/2}(\mathbf{x}_t -\mathbf{x}^*)\right\|^2.
\end{align}
\end{footnotesize}
On the other hand, let $\mathbf{x}=\mathbf{x}^*, \mathbf{y}=\mathbf{x}_t$ in Equation \eqref{2-1}. With the strong convexity of $f_t(\cdot)$, we attain
\begin{footnotesize}
\begin{align}\label{a-7}
f_t(\mathbf{x}_t) - f_t(\mathbf{x}^*) \leq \left\langle\mathbf{g}_t, \mathbf{x}_t -\mathbf{x}^*\right\rangle - \frac{\sigma}{2}\|\mathbf{x}_t -\mathbf{x}^*\|^2.
\end{align}
\end{footnotesize}
Therefore, from definition of the regret (i.e., Equation \eqref{eq1}), and Equation \eqref{a-7}, we obtain the following
\begin{footnotesize}
\begin{align}\label{a-8}
&R(T) = \sum_{t=1}^T f_t(\mathbf{x}_t) - \min_{\mathbf{x}\in\mathcal{F}}\sum_{t=1}^T f_t(\mathbf{x}) =\sum_{t=1}^T f_t(\mathbf{x}_t) - \sum_{t=1}^Tf_t(\mathbf{x}^*) \nonumber \\
&\leq \sum_{t=1}^T \left\langle\mathbf{g}_t, \mathbf{x}_t -\mathbf{x}^*\right\rangle - \frac{\sigma}{2}\sum_{t=1}^T \|\mathbf{x}_t -\mathbf{x}^*\|^2.
\end{align}
\end{footnotesize}
In addition, plugging Equation \eqref{a-6} into Equation \eqref{a-8}, we can attain the following
\begin{footnotesize}
\begin{align}\label{a-9}
&R(T) \leq \sum_{t=1}^T\frac{\left[\left\|\hat{S}_t^{1/2}\left(\mathbf{x}_t -\mathbf{x}^* \right)\right\|^2 - \left\|\hat{S}_t^{1/2}\left(\mathbf{x}_{t+1}-\mathbf{x}^*\right)\right\|^2\right]}{2\alpha_t(1-\beta_{1t})} \nonumber \\
&+ \sum_{t=1}^T\frac{\alpha_t}{2(1-\beta_{1t})}\left\|\hat{S}_t^{-1/2}\mathbf{m}_t\right\|^2 + \sum_{t=1}^T\frac{\alpha_t\beta_{1t}}{2(1-\beta_{1t})}\left\|\hat{S}_t^{-1/2}\mathbf{m}_{t-1}\right\|^2 \nonumber \\
&+ \sum_{t=1}^T\frac{\beta_{1t}}{2\alpha_t(1-\beta_{1t})}\left\|\hat{S}_t^{1/2}(\mathbf{x}_t -\mathbf{x}^*)\right\|^2 - \frac{\sigma}{2}\sum_{t=1}^T \|\mathbf{x}_t -\mathbf{x}^*\|^2.
\end{align}
\end{footnotesize}
Furthermore, since $0\leq s_{t-1,i}\leq s_{t,i}, 0\leq\alpha_t\leq\alpha_{t-1}, 0\leq\beta_{1t}\leq\beta_{1}<1$,  by Equation \eqref{a-9}, we have
\begin{footnotesize}
\begin{align}\label{a-10}
&R(T) \leq \underbrace{\sum_{t=1}^T\frac{\left[\left\|\hat{S}_t^{1/2}\left(\mathbf{x}_t -\mathbf{x}^* \right)\right\|^2 - \left\|\hat{S}_t^{1/2}\left(\mathbf{x}_{t+1}-\mathbf{x}^*\right)\right\|^2\right]}{2\alpha_t(1-\beta_{1t})}} \nonumber \\
& \underbrace{- \frac{\sigma}{2}\sum_{t=1}^T \|\mathbf{x}_t -\mathbf{x}^*\|^2}_{E_1} \nonumber \\
&+ \underbrace{\sum_{t=1}^T\frac{\alpha_t}{2(1-\beta_{1t})}\left\|\hat{S}_t^{-1/2}\mathbf{m}_t\right\|^2+ \sum_{t=2}^T\frac{\beta_{1}\alpha_{t-1}}{2(1-\beta_{1})}\left\|\hat{S}_{t-1}^{-1/2}\mathbf{m}_{t-1}\right\|^2}_{E_2} \nonumber \\
&+ \underbrace{\sum_{t=1}^T\frac{\beta_{1t}}{2\alpha_t(1-\beta_{1t})}\left\|\hat{S}_t^{1/2}(\mathbf{x}_t -\mathbf{x}^*)\right\|^2}_{E_3}.
\end{align}
\end{footnotesize}
Next, we consider the upper bounds of three parts ($E_1, E_2$ and $E_3$) in Equation \eqref{a-10} respectively. For part $E_1$, we attain the following
\begin{scriptsize}
\begin{align}\label{a-11}
&E_1 =  \frac{1}{2\alpha_1(1-\beta_{1})}\left\|\hat{S}_1^{1/2}\left(\mathbf{x}_1 -\mathbf{x}^* \right)\right\|^2    \nonumber \\
&+ \sum_{t=2}^T\frac{1}{2\alpha_t(1-\beta_{1t})}\left\|\hat{S}_t^{1/2}\left(\mathbf{x}_t -\mathbf{x}^* \right)\right\|^2 \nonumber \\
&- \sum_{t=2}^T\frac{1}{2\alpha_{t-1}(1-\beta_{1(t-1)})}\left\|\hat{S}_{t-1}^{1/2}\left(\mathbf{x}_{t}-\mathbf{x}^*\right)\right\|^2 \nonumber \\
&- \frac{1}{2\alpha_{T}(1-\beta_{1T})}\left\|\hat{S}_{T}^{1/2}\left(\mathbf{x}_{T+1}-\mathbf{x}^*\right)\right\|^2 - \frac{\sigma}{2}\sum_{t=1}^T \|\mathbf{x}_t -\mathbf{x}^*\|^2 \nonumber \\
&\leq \sum_{t=2}^T\frac{1}{1-\beta_{1t}}\Bigg[\frac{\left\|\hat{S}_t^{1/2}\left(\mathbf{x}_t -\mathbf{x}^* \right)\right\|^2}{2\alpha_t} -\frac{\left\|\hat{S}_{t-1}^{1/2}\left(\mathbf{x}_{t}-\mathbf{x}^*\right)\right\|^2}{2\alpha_{t-1}}\Bigg] \nonumber \\
&+\frac{1}{2\alpha_1(1-\beta_{1})}\left\|\hat{S}_1^{1/2}\left(\mathbf{x}_1 -\mathbf{x}^* \right)\right\|^2 - \frac{\sigma}{2}\sum_{t=1}^T \|\mathbf{x}_t -\mathbf{x}^*\|^2.
\end{align}
\end{scriptsize}
Since $\alpha_t = \frac{\alpha}{t}$ and from Equation \eqref{a-11}, we have the following
\begin{scriptsize}
\begin{align}\label{a-12}
&E_1 \leq \sum_{t=2}^T\frac{1}{1-\beta_{1t}}\Bigg[\frac{t\left\|\hat{S}_t^{1/2}\left(\mathbf{x}_t -\mathbf{x}^* \right)\right\|^2}{2\alpha}-\frac{(t-1)\left\|\hat{S}_{t-1}^{1/2}\left(\mathbf{x}_{t}-\mathbf{x}^*\right)\right\|^2}{2\alpha}\Bigg] \nonumber \\
&+\frac{1}{2\alpha_1(1-\beta_{1})}\left\|\hat{S}_1^{1/2}\left(\mathbf{x}_1 -\mathbf{x}^* \right)\right\|^2 - \frac{\sigma}{2}\sum_{t=2}^T \|\mathbf{x}_t -\mathbf{x}^*\|^2 - \frac{\sigma}{2}\|\mathbf{x}_1 -\mathbf{x}^*\|^2 \nonumber \\
&= \sum_{t=2}^T\frac{1}{2\alpha(1-\beta_{1t})}\underbrace{\Bigg[t\left\|\hat{S}_t^{1/2}\left(\mathbf{x}_t -\mathbf{x}^* \right)\right\|^2}\nonumber \\
&
\underbrace{- (t-1)\left\|\hat{S}_{t-1}^{1/2}\left(\mathbf{x}_{t}-\mathbf{x}^*\right)\right\|^2 - \sigma\alpha(1-\beta_{1t})\|\mathbf{x}_t -\mathbf{x}^*\|^2\Bigg]}_{E_1^{'}} \nonumber \\
&+\underbrace{\frac{1}{2\alpha_1(1-\beta_{1})}\left\|\hat{S}_1^{1/2}\left(\mathbf{x}_1 -\mathbf{x}^* \right)\right\|^2 - \frac{\sigma}{2}\|\mathbf{x}_1 -\mathbf{x}^*\|^2}_{E_1^{''}}.
\end{align}
\end{scriptsize}
In addition, for term $E_1^{'}$ of Equation \eqref{a-12}, and from Equation \eqref{2-5}, we can obtain
\begin{footnotesize}
\begin{align}\label{a-13}
E_1^{'}&= \sum_{i=1}^{n}(x_{t,i}-x^*_{,i})^2\Bigg[ t\left(\hat{s}_{t,i}+\frac{\delta}{t}\right)^{1/2}
\nonumber \\
&- (t-1)\left(\hat{s}_{t-1,i}+\frac{\delta}{t-1}\right)^{1/2} - \sigma\alpha(1-\beta_{1t})\Bigg],
\end{align}
\end{footnotesize}
where $i\in\{1,\ldots,n\}$, and $n$ is the dimension of decision vectors.
Moreover, since  $(a+b)^{1/2}\leq a^{1/2}+b^{1/2}$ and $1-\beta_1 \leq 1-\beta_{1t}$,  by Equation \eqref{c-2}, we can have the following
\begin{scriptsize}
\begin{align}\label{a-14}
&E_1^{'} \leq \sum_{i=1}^{n}(x_{t,i}-x^*_{,i})^2\Bigg[t\hat{s}_{t,i}^{1/2} - (t-1)\hat{s}_{t-1,i}^{1/2} - \sqrt{\delta(t-1)} - \sigma\alpha(1-\beta_{1t})\Bigg] \nonumber \\
&\leq \sum_{i=1}^{n}(x_{t,i}-x^*_{,i})^2\Bigg[ts_{t,i}^{1/2} - (t-1)s_{t-1,i}^{1/2} - \sqrt{\delta(t-1)}- \sigma\alpha(1-\beta_{1t})\Bigg] \nonumber \\
&\leq \sum_{i=1}^{n}(x_{t,i}-x^*_{,i})^2\Bigg[\sigma\alpha(1-\beta_1) - \sqrt{\delta(t-1)}- \sigma\alpha(1-\beta_{1t})\Bigg]\nonumber \\
&\leq 0.
\end{align}
\end{scriptsize}
Next, for term $E_1^{''}$ of Equation \eqref{a-12}, we have
\begin{footnotesize}
\begin{align}\label{a-15}
E_1^{''} &= \frac{1}{2\alpha_1(1-\beta_{1})}\left\|\hat{S}_1^{1/2}\left(\mathbf{x}_1 -\mathbf{x}^* \right)\right\|^2 - \frac{\sigma}{2}\|\mathbf{x}_1 -\mathbf{x}^*\|^2 \nonumber \\
&= \sum_{i=1}^{n}\left[\frac{s_1+\delta-\sigma\alpha_1(1-\beta_1)}{2\alpha_1(1-\beta_1)}\right](x_{1,i}-x^*_{,i})^2.
\end{align}
\end{footnotesize}
By Equation \eqref{c-2}, $s_1 - \sigma\alpha_1(1-\beta_1)\leq 0$ when $t=1$, and from Equation \eqref{a-15}, we obtain the following
\begin{footnotesize}
\begin{align}\label{a-16}
E_1^{''} &\leq \sum_{i=1}^{n}\frac{\delta}{2\alpha_1(1-\beta_1)}(x_{1,i}-x^*_{,i})^2.
\end{align}
\end{footnotesize}
Moreover, from Assumption \ref{ass1}, we have that $x_{1,i}-x^*_{,i}\leq D_{\infty}$. Then, combining Equations \eqref{a-12}, \eqref{a-14} and \eqref{a-16}, we finally attain
\begin{footnotesize}
\begin{align}\label{a-17}
E_1 \leq \sum_{i=1}^{n}\frac{\delta D_{\infty}^2}{2\alpha_1(1-\beta_1)} = \frac{n\delta D_{\infty}^2}{2\alpha(1-\beta_1)}.
\end{align}
\end{footnotesize}
Therefore, we have obtained the upper bound of part $E_1$. Then we consider part $E_2$ of the remaining two parts in Equation \eqref{a-10}. Since $\beta_{1t}\leq\beta_1$, we have the following
\begin{scriptsize}
\begin{align}\label{a-18}
E_2 &\leq \sum_{t=1}^T\frac{\alpha_t}{2(1-\beta_{1})}\left\|\hat{S}_t^{-1/2}\mathbf{m}_t\right\|^2+ \sum_{t=2}^T\frac{\beta_{1}\alpha_{t-1}}{2(1-\beta_{1})}\left\|\hat{S}_{t-1}^{-1/2}\mathbf{m}_{t-1}\right\|^2.
\end{align}
\end{scriptsize}
For Equation \eqref{a-18}, we first consider the term $\sum_{t=1}^T\alpha_t\left\|\hat{S}_t^{-1/2}\mathbf{m}_t\right\|^2$,  we obtain
\begin{footnotesize}
\begin{align}\label{a-19}
&\sum_{t=1}^T\alpha_t\left\|\hat{S}_t^{-1/2}\mathbf{m}_t\right\|^2= \sum_{t=1}^{T-1}\alpha_t\left\|\hat{S}_t^{-1/2}\mathbf{m}_t\right\|^2 + \alpha_T\left\|\hat{S}_T^{-1/2}\mathbf{m}_T\right\|^2 \nonumber \\
&\leq \sum_{t=1}^{T-1}\alpha_t\left\|\hat{S}_t^{-1/2}\mathbf{m}_t\right\|^2 + \underbrace{\alpha_T\sum_{i=1}^{n}\frac{m_{T,i}}{s_{T,i}+\frac{\delta}{T}}}_{E_2^{'}}.
\end{align}
\end{footnotesize}
Moreover, assuming $\frac{g_{t,i} - m_{t,i}}{g_{t,i}} = \varpi_t$ where $\varpi_t\in(0,1)$ and $i\in\{1,\ldots,n\}$, and let $\varpi = \min\{\varpi_1, \ldots, \varpi_t\}$. Furthermore, applying the recursive algorithm to \eqref{2-2} and \eqref{2-3}, we have that
\begin{footnotesize}
\begin{align}\label{a-19-1}
E_2^{'}&= \alpha\sum_{i=1}^{n}\frac{\left(\sum_{j=1}^{T}(1-\beta_{1j})\prod_{k=1}^{T-j}\beta_{1(T-k+1)}g_{j,i}\right)^2}{T\sum_{j=1}^{T}(1-\beta_{2j})\prod_{k=1}^{T-j}\beta_{2(T-k+1)}(g_{j,i}-m_{j,i})^2+\delta} \nonumber \\
&\leq\alpha\sum_{i=1}^{n}\frac{\left(\sum_{j=1}^{T}(1-\beta_{1j})\prod_{k=1}^{T-j}\beta_{1(T-k+1)}g_{j,i}\right)^2}{\varpi^2 T\sum_{j=1}^{T}(1-\beta_{2j})\prod_{k=1}^{T-j}\beta_{2(T-k+1)}g_{j,i}^2+\delta}.
\end{align}
\end{footnotesize}
From Equation \eqref{a-19-1} and $\beta_{1t}\in (0,1]$, $E_2^{'}$ can be further bounded as
\begin{footnotesize}
\begin{align}\label{a-20}
&E_2^{'} \leq \alpha\sum_{i=1}^{n}\frac{\left(\sum_{j=1}^{T}\prod_{k=1}^{T-j}\beta_{1(T-k+1)}g_{j,i}\right)^2}{\varpi^2 T\sum_{j=1}^{T}(1-\beta_{2j})\prod_{k=1}^{T-j}\beta_{2(T-k+1)}g_{j,i}^2+\delta}.
\end{align}
\end{footnotesize}
Furthermore, according to Cauchy-Schwarz inequality, i.e., $\left(\sum_{k=1}^n \langle a_k,b_k\rangle\right)^2\leq\left(\sum_{k=1}^n a_k^2\right)\left(\sum_{k=1}^n b_k^2\right)$, and from Equation \eqref{a-20}, we attain the following
\begin{scriptsize}
\begin{align}\label{a-20-1}
&E_2^{'}\leq\alpha\sum_{i=1}^{n}\frac{\left(\sum_{j=1}^{T}\prod_{k=1}^{T-j}\beta_{1(T-k+1)}\right)\left(\sum_{j=1}^{T}\prod_{k=1}^{T-j}\beta_{1(T-k+1)}g_{j,i}^2\right)}{\varpi^2 T\sum_{j=1}^{T}(1-\beta_{2j})\prod_{k=1}^{T-j}\beta_{2(T-k+1)}g_{j,i}^2+\delta} \nonumber \\
&\leq \alpha\sum_{i=1}^{n}\frac{\left(\sum_{j=1}^{T}\beta_1^{T-j}\right)\left(\sum_{j=1}^{T}\prod_{k=1}^{T-j}\beta_{1(T-k+1)}g_{j,i}^2\right)}{\varpi^2 T\sum_{j=1}^{T}(1-\beta_{2j})\prod_{k=1}^{T-j}\beta_{2(T-k+1)}g_{j,i}^2+\delta}.
\end{align}
\end{scriptsize}
Since $\beta_{1t}\leq\beta_1$ and from Equation \eqref{a-20-1}, we can further attain the following bound for $E_2^{'}$:
\begin{footnotesize}
\begin{align}\label{a-21}
E_2^{'} &\leq \frac{\alpha}{\varpi^2(1-\beta_1)}\sum_{i=1}^{n}\frac{\sum_{j=1}^{T}\prod_{k=1}^{T-j}\beta_{1(T-k+1)}g_{j,i}^2}{ T\sum_{j=1}^{T}(1-\beta_{2j})\prod_{k=1}^{T-j}\beta_{2(T-k+1)}g_{j,i}^2+\delta} \nonumber \\
&\leq \frac{\alpha}{\varpi^2(1-\beta_1)}\sum_{i=1}^{n}\frac{\sum_{j=1}^{T}\beta_{1}^{T-j}g_{j,i}^2}{ T\sum_{j=1}^{T}(1-\beta_{2j})\prod_{k=1}^{T-j}\beta_{2(T-k+1)}g_{j,i}^2+\delta}.
\end{align}
\end{footnotesize}
By Equation \eqref{c-1}, we have the following
\begin{footnotesize}
\begin{align}\label{a-22}
E_2^{'} &\leq \frac{\alpha\zeta}{\varpi^2(1-\beta_1)}\sum_{i=1}^{n}\frac{\sum_{j=1}^{T}\beta_{1}^{T-j}g_{j,i}^2}{ \sum_{j=1}^{T}g_{j,i}^2+\zeta\delta} \nonumber \\
&\leq \frac{\alpha\zeta}{\varpi^2(1-\beta_1)}\sum_{i=1}^{n}\sum_{j=1}^{T}\beta_{1}^{T-j}\frac{g_{j,i}^2}{\sum_{k=1}^{j}g_{k,i}^2+\zeta\delta}.
\end{align}
\end{footnotesize}
Moreover, plugging Equation \eqref{a-22} into Equation \eqref{a-19}, and applying recursive algorithm, we attain
\begin{scriptsize}
\begin{align}\label{a-23}
&\sum_{t=1}^T\alpha_t\left\|\hat{S}_t^{-1/2}\mathbf{m}_t\right\|^2 \leq \sum_{t=1}^{T-1}\alpha_t\left\|\hat{S}_t^{-1/2}\mathbf{m}_t\right\|^2 \nonumber \\
&+ \frac{\alpha\zeta}{\varpi^2(1-\beta_1)}\sum_{i=1}^{n}\sum_{j=1}^{T}\beta_{1}^{T-j}\frac{g_{j,i}^2}{\sum_{k=1}^{j}g_{j,i}^2+\zeta\delta} \nonumber \\
&\leq \frac{\alpha\zeta}{\varpi^2(1-\beta_1)}\sum_{i=1}^{n}\sum_{t=1}^{T}\sum_{j=1}^{t}\beta_{1}^{t-j}\frac{g_{j,i}^2}{\sum_{k=1}^{j}g_{k,i}^2+\zeta\delta} \nonumber \\
&\leq \frac{\alpha\zeta}{\varpi^2(1-\beta_1)}\sum_{i=1}^{n}\sum_{j=1}^{T}\sum_{l=0}^{T-j}\beta_{1}^{l}\frac{g_{j,i}^2}{\sum_{k=1}^{j}g_{k,i}^2+\zeta\delta} \nonumber \\
&\leq \frac{\alpha\zeta}{\varpi^2(1-\beta_1)}\sum_{i=1}^{n}\sum_{j=1}^{T}\frac{\sum_{l=0}^{T-j}\beta_{1}^{l}g_{j,i}^2}{\sum_{k=1}^{j}g_{k,i}^2+\zeta\delta}  \nonumber \\
&\leq \frac{\alpha\zeta}{\varpi^2(1-\beta_1)^2}\sum_{i=1}^{n}\sum_{j=1}^{T}\frac{g_{j,i}^2}{\sum_{k=1}^{j}g_{k,i}^2+\zeta\delta}.
\end{align}
\end{scriptsize}
If let $\Psi= \frac{g_{j,i}^2}{\sum_{k=1}^{j}g_{j,i}^2+\zeta\delta}$, $\omega_j=\sum_{k=1}^{j}g_{k,i}^2+\zeta\delta$, and $\omega_0=\zeta\delta$, we obtain
\begin{footnotesize}
\begin{align}\label{a-24}
\Psi = \frac{\omega_j - \omega_{j-1}}{\omega_j}.
\end{align}
\end{footnotesize}
In addition, for any $a\geq b> 0$, the inequality $1+x\leq e^x$ implies that
\begin{footnotesize}
\begin{align}\label{a-25}
\frac{a-b}{a}\leq \log \frac{a}{b}.
\end{align}
\end{footnotesize}
Therefore, by Equation \eqref{a-25}, Equation \eqref{a-24} has the following bound:
\begin{footnotesize}
\begin{align}\label{a-25-1}
\Psi\leq\log\frac{\omega_{j}}{\omega_{j-1}}.
\end{align}
\end{footnotesize}
Plugging Equation \eqref{a-25-1} into Equation \eqref{a-23}, we have
\begin{scriptsize}
\begin{align}\label{a-25-2}
&\sum_{t=1}^T\alpha_t\left\|\hat{S}_t^{-1/2}\mathbf{m}_t\right\|^2 \leq \frac{\alpha\zeta}{\varpi^2(1-\beta_1)^2}\sum_{i=1}^{n}\sum_{j=1}^{T}\log\frac{\omega_{j}}{\omega_{j-1}} \nonumber \\
&\leq \frac{\alpha\zeta}{\varpi^2(1-\beta_1)^2}\sum_{i=1}^{n}\log\frac{\omega_{T}}{\omega_{0}} \leq \frac{\alpha\zeta}{\varpi^2(1-\beta_1)^2}\sum_{i=1}^{n}\log\left(\frac{\sum_{k=1}^{T}g_{k,i}^2}{\zeta\delta}+1\right).
\end{align}
\end{scriptsize}
From Equations \eqref{a-18} and \eqref{a-25-2}, $E_2$ can be further bounded as
\begin{footnotesize}
\begin{align}\label{a-27}
E_2 &\leq \frac{1}{1-\beta_{1}}\sum_{t=1}^T\alpha_t\left\|\hat{S}_t^{-1/2}\mathbf{m}_t\right\|^2 \nonumber \\
&\leq \frac{\alpha\zeta}{\varpi^2(1-\beta_1)^3}\sum_{i=1}^{n}\log\left(\frac{\sum_{k=1}^{T}g_{k,i}^2}{\zeta\delta}+1\right) \nonumber \\
&\leq \frac{\alpha\zeta}{\varpi^2(1-\beta_1)^3}\sum_{i=1}^{n}\log\left(\frac{1}{\zeta\delta}\|g_{1:T,i}\|^2+1\right).
\end{align}
\end{footnotesize}

Next, we consider the last term $E_3$ in Equation \eqref{a-10}. From the definition of $\hat{S}_t$ and Assumption \ref{ass2}, we obtain the following
\begin{footnotesize}
\begin{align}\label{a-28}
E_3 &\leq \sum_{t=1}^T\frac{t\beta_{1t}}{2\alpha(1-\beta_{1t})}\left\|\left(\mathbf{s}_t+\frac{\delta}{t}\right)^{1/2}(\mathbf{x}_t -\mathbf{x}^*)\right\|^2 \nonumber \\
&\leq \sum_{i=1}^n\sum_{t=1}^T\frac{t\beta_{1t}}{2\alpha(1-\beta_{1t})}\left\|\left(s_{t,i}+\frac{\delta}{t}\right)^{1/2}(x_{t,i} -x_{,i}^*)\right\|^2 \nonumber \\
&\leq \frac{D_{\infty}^2(G_{\infty}+\delta)}{2\alpha}\sum_{i=1}^n\sum_{t=1}^T\frac{\beta_{1t}}{1-\beta_{1t}}t.
\end{align}
\end{footnotesize}
Finally, combining Equations \eqref{a-10}, \eqref{a-17}, \eqref{a-27} and \eqref{a-28}, we obtain the upper bound of $R(T)$ as follows
\begin{footnotesize}
\begin{align}\label{a-29}
R(T) \leq &\frac{n\delta D_{\infty}^2}{2\alpha(1-\beta_1)}+\frac{D_{\infty}^2(G_{\infty}+\delta)}{2\alpha}\sum_{i=1}^n\sum_{t=1}^T\frac{\beta_{1t}}{1-\beta_{1t}}t  \nonumber \\
&+ \frac{\alpha\zeta}{\varpi^2(1-\beta_1)^3}\sum_{i=1}^{n}\log\left(\frac{1}{\zeta\delta}\|g_{1:T,i}\|^2+1\right).
\end{align}
\end{footnotesize}
Therefore, the proof of Theorem \ref{thm1} is completed.
$\hfill\blacksquare$

\textbf{Corollary 1}
Let $\beta_{1t}=\beta_1\lambda^t$, where $\lambda\in(0,1)$ in Theorem \ref{thm1}. Then we have the following upper bound of the regret
\begin{footnotesize}
\begin{align*}
R(T) &\leq \frac{n\delta D_{\infty}^2}{2\alpha(1-\beta_1)} +\frac{n\beta_{1}\lambda D_{\infty}^2(G_{\infty}+\delta)}{2\alpha(1-\beta_{1})(1-\lambda)^2} \nonumber \\
& + \frac{\alpha\zeta}{\varpi^2(1-\beta_1)^3}\sum_{i=1}^{n}\log\left(\frac{1}{\zeta\delta}\|g_{1:T,i}\|^2+1\right).
\end{align*}
\end{footnotesize}

\paragraph{\emph{Proof.}}
Since $\beta_{1t}=\beta_1\lambda^t$, Equation \eqref{a-28} can be further bounded as follows
\begin{footnotesize}
\begin{align}\label{a-30}
E_3 &\leq \frac{\beta_{1}D_{\infty}^2(G_{\infty}+\delta)}{2\alpha(1-\beta_{1})}\sum_{i=1}^n\sum_{t=1}^T t\lambda^t \nonumber \\
&\leq \frac{\beta_{1}D_{\infty}^2(G_{\infty}+\delta)}{2\alpha(1-\beta_{1})}\sum_{i=1}^n\left[\frac{\lambda(1-\lambda^T)}{(1-\lambda)^2}-\frac{T\lambda^{T+1}}{1-\lambda}\right] \nonumber \\
&\leq \frac{\beta_{1}D_{\infty}^2(G_{\infty}+\delta)}{2\alpha(1-\beta_{1})}\sum_{i=1}^n\frac{\lambda}{(1-\lambda)^2} = \frac{n\beta_{1}\lambda D_{\infty}^2(G_{\infty}+\delta)}{2\alpha(1-\beta_{1})(1-\lambda)^2}.
\end{align}
\end{footnotesize}
In addition, plugging Equation \eqref{a-30} into Equation \eqref{a-29}, we further attain the upper bound of $R(T)$ as follows
\begin{footnotesize}
\begin{align}\label{a-31}
R(T) &\leq \frac{n\delta D_{\infty}^2}{2\alpha(1-\beta_1)} +\frac{n\beta_{1}\lambda D_{\infty}^2(G_{\infty}+\delta)}{2\alpha(1-\beta_{1})(1-\lambda)^2} \nonumber \\
& + \frac{\alpha\zeta}{\varpi^2(1-\beta_1)^3}\sum_{i=1}^{n}\log\left(\frac{1}{\zeta\delta}\|g_{1:T,i}\|^2+1\right).
\end{align}
\end{footnotesize}
Therefore, the proof of Corollary \ref{cor1} is completed.
$\hfill\blacksquare$

\section{Details on Conditions 3 and 4.} \label{appendixb}
The contribution of condition 3 and condition 4 is to ensure the convergence of the proposed algorithm, which are common conditions in many second-order momentum algorithms, such as AMSGrad, SAdam. However, in the actual training task, the parameters in these conditions are not explicitly involved, so they are only used in the proof process.

For condition 3, the parameter $\zeta$ only needs to satisfy that
\begin{footnotesize}
\begin{align}
t\sum_{j=1}^t\prod_{k=1}^{t-j}\beta_{2(t-k+1)}(1-\beta_{2j})g_{j,i}^2 \geq \frac{1}{\zeta}\sum_{j=1}^t g_{j,i}^2.  \nonumber
\end{align}
\end{footnotesize}
From the above inequation, we have that
\begin{footnotesize}
\begin{align}
\frac{1}{\zeta}\geq t\prod_{k=1}^{t-j}\beta_{2(t-k+1)}(1-\beta_{2j}) > t. \nonumber
\end{align}
\end{footnotesize}
Therefore, the value of parameter $\zeta$ is easy to choose.

For condition 4, from the definition of $\mathbf{s}_t$:
\begin{footnotesize}
\begin{align}
\mathbf{s}_t=\beta_{2t}\mathbf{s}_{t-1}+(1-\beta_{2t})(\mathbf{g}_t-\mathbf{m}_t)^2,\nonumber
\end{align}
\end{footnotesize}
and applying the recursive algorithm for the above equation, we have:
\begin{footnotesize}
\begin{align}
\mathbf{s}_t &= (1-\beta_{2t})(\mathbf{g}_t-\mathbf{m}_t)^2 +\beta_{2(t-1)}(1-\beta_{2(t-1)})(\mathbf{g}_{t-1}-\mathbf{m}_{t-1})^2 \nonumber \\
&+ \ldots + \beta_{20}^{T-1}(1-\beta_{20})(\mathbf{g}_{0}-\mathbf{m}_{0})^2,\nonumber
\end{align}
\end{footnotesize}
and
\begin{footnotesize}
\begin{align}
\mathbf{s}_{t-1} &=(1-\beta_{2t})(\mathbf{g}_t-\mathbf{m}_t)^2 +\beta_{2(t-1)}(1-\beta_{2(t-1)})(\mathbf{g}_{t-1}-\mathbf{m}_{t-1})^2 \nonumber \\
&+ \ldots + \beta_{20}^{T-1}(1-\beta_{20})(\mathbf{g}_{0}-\mathbf{m}_{0})^2,
\end{align}
\end{footnotesize}
thus we further obtain:
\begin{footnotesize}
\begin{align}
&\frac{t}{\alpha}\mathbf{s}_t-\frac{t-1}{\alpha}\mathbf{s}_{t-1} \leq \frac{t}{\alpha}(1-\beta_{2t})(\mathbf{g}_t-\mathbf{m}_t)^2 + \frac{1}{\alpha}\mathbf{s}_{t-1} \nonumber \\
&\leq \frac{T}{\alpha}G_{\infty}^2 + \frac{1}{\alpha}(1-\beta_{2}^{T-1})G_{\infty}^2 = \frac{1}{\alpha}\left(T+1-\beta_{2}^{T-1}\right)G_{\infty}^2. \nonumber
\end{align}
\end{footnotesize}
Therefore, the parameter $\sigma$ only needs to be selected in the following way to make condition 3 true:
\begin{footnotesize}
\begin{align}
\sigma \geq \frac{1}{\alpha(1-\beta_1)}\left(T+1-\beta_{2}^{T-1}\right)G_{\infty}^2. \nonumber
\end{align}
\end{footnotesize}
In summary, the selection of parameters in conditions 3 and 4 is not difficult.

\section{Experiments on CIFAR-100.}
\begin{figure*}
  \begin{minipage}[t]{0.31\linewidth}
    \centering
		\includegraphics[scale=0.22]{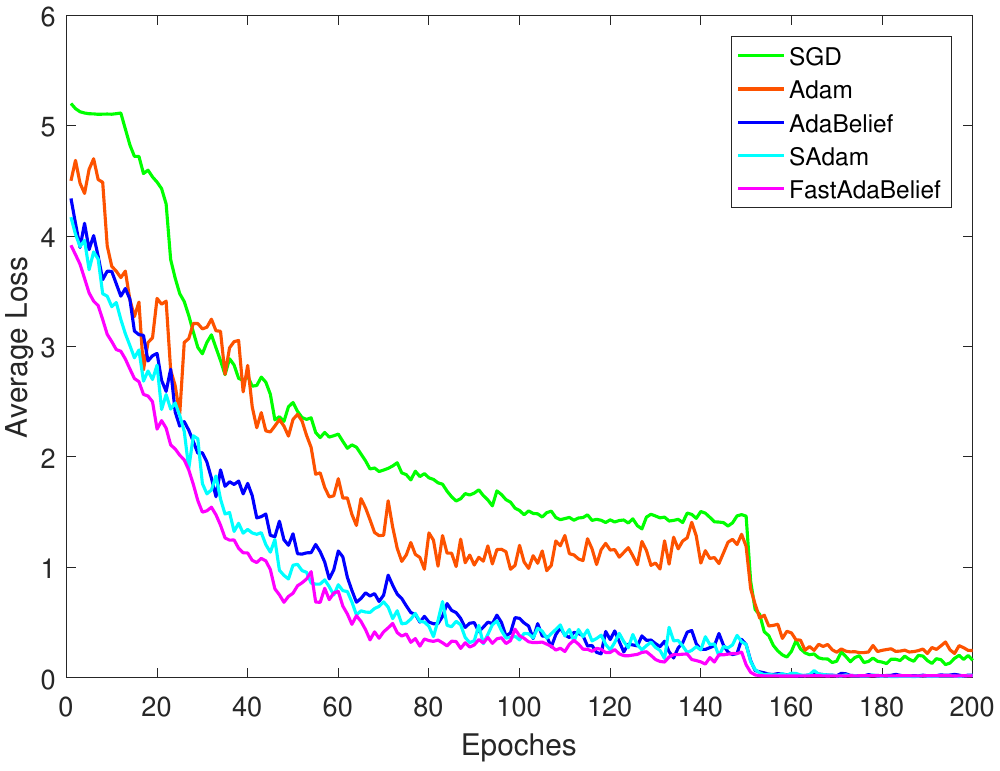}
	    \caption{Comparison of the average loss of SGD, Adam, AdaBelief, SAdam, and FastAdaBelief on CIFAR-100. FastAdaBelief converges faster than other algorithms.}
	    \label{fig6}
  \end{minipage}
  \begin{minipage}[t]{0.31\linewidth}
    \centering
		\includegraphics[scale=0.22]{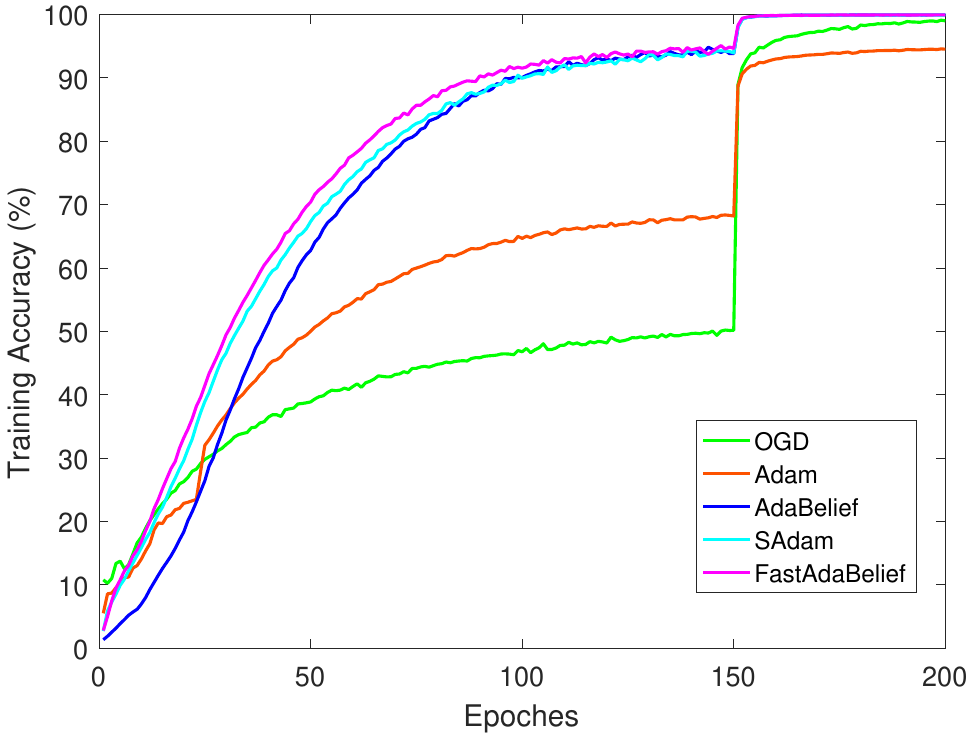}
	    \caption{Comparison of the training accuracy of SGD, Adam, AdaBelief, SAdam, and FastAdaBelief on CIFAR-100. FastAdaBelief outperforms other algorithms on training accuracy.}
	    \label{fig7}
  \end{minipage}
  \begin{minipage}[t]{0.31\linewidth}
    \centering
		\includegraphics[scale=0.25]{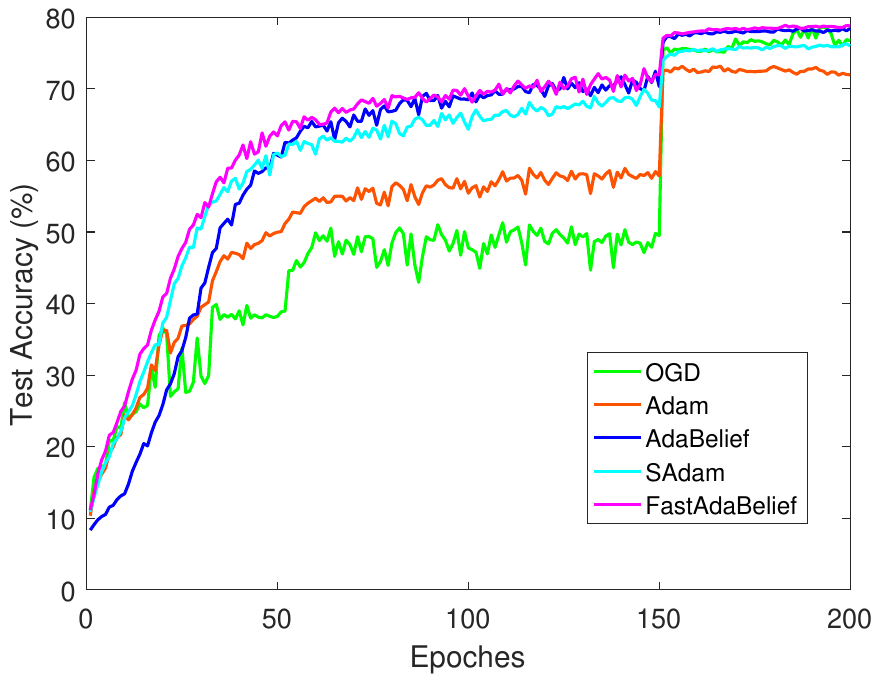}
	    \caption{Comparison of the performance of SGD, Adam, AdaBelief, SAdam, and FastAdaBelief on CIFAR-100. FastAdaBelief outperforms other algorithms on test accuracy.}
	    \label{fig8}
  \end{minipage}
\end{figure*}

We experiment 5 times with a DenseNet-121 model on CIFAR-100 classication task. The parameters are set as same as in Section \ref{Experiments}. The results shown in Figure \ref{fig6} demonstrate that the proposed algorithm converges faster than other algorithms. In Figure \ref{fig7},  the top-1 training accuracy of 5 independent runs of the proposed algorithm is the highest in the algorithms. In addition, Figure \ref{fig8} shows that  the top-1 test accuracy of the proposed algorithm better than other algorithms.

\section{Convergence analysis of FastAdaBelief in convex conditions}\label{app2}
The regret bound analysis for the proposed algorithm when loss functions are convex are present as below:
\textbf{Proof.} Reviewing the forms of the proposed algorithm and AdaBelief that the difference is their stepsizes. The stepsizes of the proposed algorithm and AdaBelief are shown as below:
\begin{footnotesize}
\begin{align}
 \Delta_t(FastAdaBelief)=\frac{\alpha_t \mathbf{m}_t}{\hat{S}_t}=\frac{\alpha \mathbf{m}_t}{t\mathbf{s}_t+\delta},\quad \quad (\alpha_t=1/t)
\end{align}
\end{footnotesize}
and
\begin{footnotesize}
\begin{align}
\Delta_t(AdaBelief)= \frac{\alpha_t\mathbf{m}_t}{\sqrt{\hat{S}_t}} = \frac{\alpha\mathbf{m}_t}{\sqrt{t(\mathbf{s}_t+\epsilon)}},\quad \quad (\alpha_t=1/\sqrt{t})
\end{align}
\end{footnotesize}
where $\mathbf{s}_t = \beta_{2t}\mathbf{s}_{t-1}+(1-\beta_{2t})(\mathbf{g}_t-\mathbf{s}_t)^2$. Note that $\epsilon$ is a very small positive term to keep the denominator from going to zero, which is ignored in the convergence analysis of AdaBelief. Therefore, $\delta$, which has the same effect as $\epsilon$, will also be ignored in the convergence analysis of the proposed algorithm.

To attain the regret bound for the proposed algorithm when the loss functions are convex, we first consider the bound of the following term.
\begin{footnotesize}
\begin{align}\label{1}
\left\|\mathbf{s}^{\frac{1}{2}}\left(\mathbf{x}_{t+1}-\mathbf{x}^*\right)\right\| &\leq \left\|\mathbf{s}_t^{\frac{1}{2}}\left(\mathbf{x}_{t}-\mathbf{x}^*\right)\right\|^2 + \alpha_t^2\left\|\mathbf{s}_t^{-\frac{1}{2}}\mathbf{m}_t\right\|^2 \nonumber \\ &-2\alpha_t\left\langle\beta_{1t}\mathbf{m}_{t-1}+(1-\beta_{1t})\mathbf{g}_t, \mathbf{x}_{t}-\mathbf{x}^*\right\rangle.
\end{align}
\end{footnotesize}
Rearranging inequation \eqref{1}, we have:
\begin{footnotesize}
\begin{align}\label{2}
&\left\langle\mathbf{g}_t, \mathbf{x}_{t}-\mathbf{x}^*\right\rangle \nonumber \\
&\leq \frac{1}{2\alpha_t(1-\beta_{1t})}\left(\left\|\mathbf{s}_t^{1/2}(\mathbf{x}_t-\mathbf{x}^*)\right\|^2-\left\|\mathbf{s}_t^{1/2}(\mathbf{x}_{t+1}-\mathbf{x}^*)\right\|^2 \right) \nonumber \\
&+\frac{1}{2(1-\beta_{1t})}\left\|\mathbf{s}_t^{-1/2}\mathbf{m}_t\right\|^2 + \frac{\beta_{1t}\alpha_t}{2(1-\beta_{1t})}\left\|\mathbf{s}_t^{-1/2}\mathbf{m}_{t-1}\right\|^2 \nonumber \\
&+ \frac{\beta_{1t}}{2\alpha_t(1-\beta_{1t})}\left\|\mathbf{s}_t^{1/2}\left(\mathbf{m}_{t}-\mathbf{x}^*\right)\right\|^2.
\end{align}
\end{footnotesize}
By the convexity of function, we attain:
\begin{scriptsize}
\begin{align}\label{3}
&\sum_{t=1}^T \left[f_t(\mathbf{x}_t)-f_t(\mathbf{x}^*)\right] \leq \sum_{t=1}^T \left\langle\mathbf{g}_t, \mathbf{x}_{t}-\mathbf{x}^*\right\rangle \nonumber \\
&\leq \frac{1}{2(1-\beta_1)}\frac{\big\|\mathbf{s}^{1/2}(\mathbf{x}_1-\mathbf{x}^*)\big\|^2}{\alpha_1}\nonumber \\
& + \frac{1}{2(1-\beta_1)}\sum_{t=2}^T\big\|\mathbf{x}_t-\mathbf{x}^*\big\|^2\left[\frac{\mathbf{s}_t^{1/2}}{\alpha_t}-\frac{\mathbf{s}_{t-1}^{1/2}}{\alpha_{t-1}}\right] \nonumber \\
&+\frac{1+\beta_1}{2(1-\beta_1)}\sum_{t=1}^T\alpha_t\big\|\mathbf{s}_t^{-1/2}\mathbf{m}_t\big\|^2 + \frac{1}{2(1-\beta_1)}\sum_{t=1}^T\frac{\beta_{1t}}{\alpha_t}\big\|\mathbf{s}_t^{1/2}(\mathbf{x}_t-\mathbf{x}^*)\big\|^2.
\end{align}
\end{scriptsize}
Assuming $0<c<\|\mathbf{s}_t\|$, the term $\sum_{t=1}^T\alpha_t\big\|\mathbf{s}_t^{-1/2}\mathbf{m}_t\big\|^2$ in inequation \eqref{3} can be bounded as follows:
\begin{scriptsize}
\begin{align}\label{4}
&\sum_{t=1}^T\alpha_t\big\|\mathbf{s}_t^{-1/2}\mathbf{m}_t\big\|^2\leq \sum_{t=1}^{T-1}\alpha_t\big\|\mathbf{s}_t^{-1/2}\mathbf{m}_t\big\|^2 + \frac{\alpha}{cT}\|\mathbf{m}_T\|^2 \nonumber \\
&\leq \sum_{t=1}^{T-1}\alpha_t\big\|\mathbf{s}_t^{-1/2}\mathbf{m}_t\big\|^2 + \frac{\alpha}{cT}\sum_{i=1}^d\left(\sum_{j=1}^T(1-\beta_{1,j})g_{j,i}\prod_{k=1}^{T-j}\beta_{1,T-k+1}\right)^2 \nonumber \\
&\leq \frac{\alpha}{c(1-\beta_1)^2}\sum_{i=1}^d\left\|g_{1:T,i}^2\right\|\frac{1}{t}\leq \frac{\alpha(1+\log T)}{c(1-\beta_1)^2}\sum_{i=1}^d\left\|g_{1:T,i}^2\right\|.
\end{align}
\end{scriptsize}
Applying inequation \eqref{4} into inequation \eqref{3}, we obtain:
\begin{scriptsize}
\begin{align}\label{5}
R(T)&=\sum_{t=1}^T \left[f_t(\mathbf{x}_t)-f_t(\mathbf{x}^*)\right] \leq \frac{D_{\infty}^2 T}{2\alpha(1-\beta_1)}\sum_{i=1}^d s_{T,i}^{1/2} \nonumber \\
&+ \frac{(1-\beta)\alpha\log T}{2c(1-\beta_1)^3}\sum_{i=1}^d\left\|g_{1:T,i}^2\right\|+\frac{D_{\infty}^2}{2(1-\beta_1)}\sum_{t=1}^T\sum_{i=1}^d\frac{\beta_{1t}s_{t,i}^{1/2}}{\alpha_t}.
\end{align}
\end{scriptsize}
Therefore, the proof of the regret bound of the proposed algorithm when loss functions are convex is completed.
$\hfill\blacksquare$ \\

\begin{IEEEbiography}[{\includegraphics[width=1in,height=1.25in,clip,keepaspectratio]{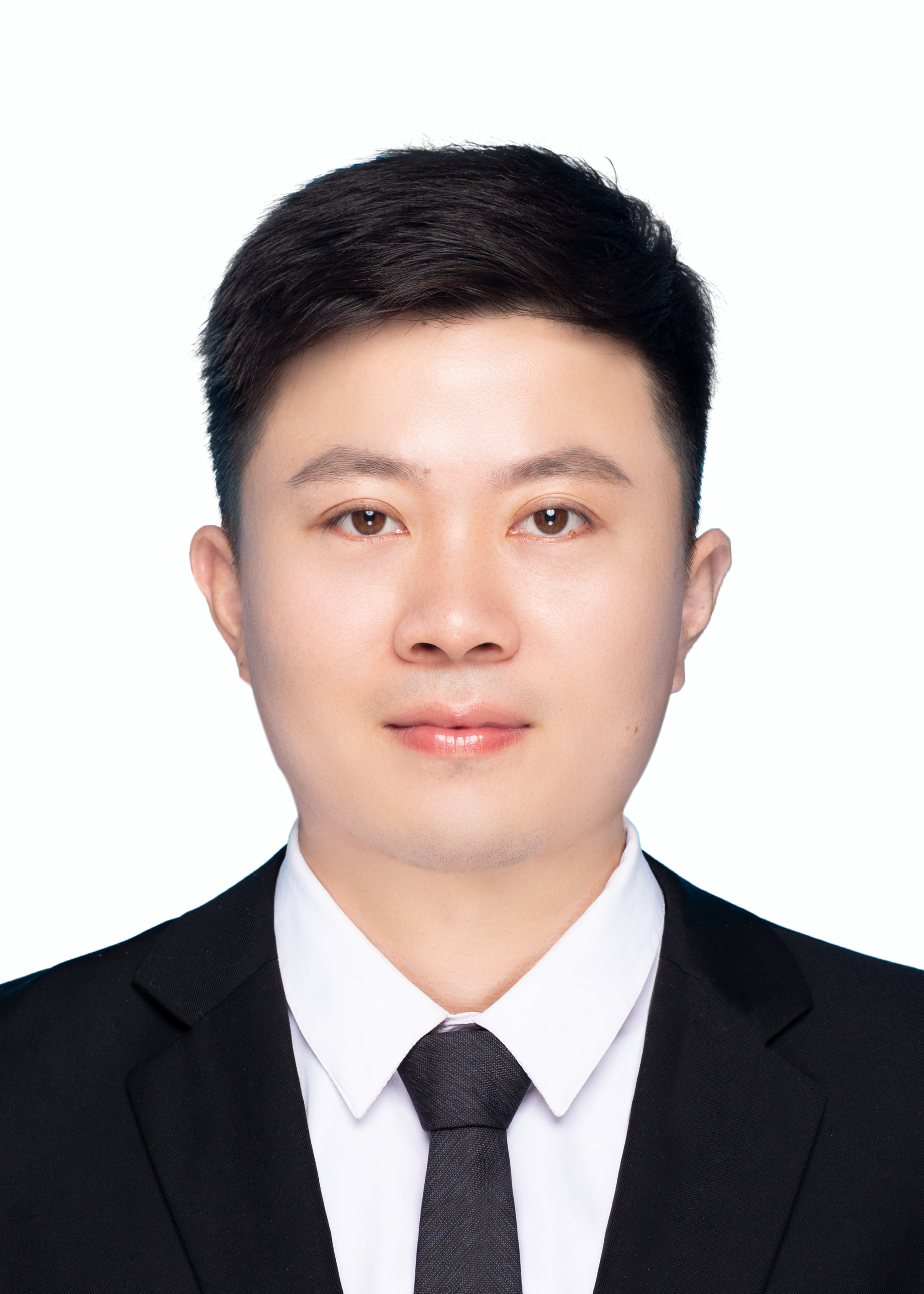}}]{Yangfan Zhou}
is currently pursuing the Ph.D. degree in the School of Nano-Tech and Nano-Bionics, University of Science and Technology of China. His current research interests are focused theoretical and algorithmic issues related to on large-scale optimization, stochastic optimization, convex online optimization, and their applications to deep learning, meta learning, and networking. He is currently a reviewer for the IEEE Transactions on Neural Networks and Learning Systems.
\end{IEEEbiography}

\begin{IEEEbiography}[{\includegraphics[width=1in,height=1.25in,clip,keepaspectratio]{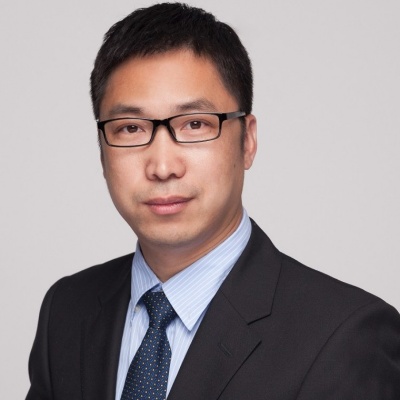}}]{Kaizhu Huang} is currently a Professor at Department of Intelligent Science, Xi'an Jiaotong-Liverpool University (XJTLU), China.  He acts as associate dean of research in School of Advanced Technology, XJTLU and is also the founding director of Suzhou Municipal Key Laboratory of Cognitive Computation and Applied Technology. Prof. Huang obtained his PhD degree from Chinese University of Hong Kong (CUHK) in 2004. He worked in Fujitsu Research Centre, CUHK, University of Bristol, National Laboratory of Pattern Recognition, Chinese Academy of Sciences from 2004 to 2012. Prof. Huang has been working in pattern recognition, machine learning, and neural information processing. He was the recipient of 2011 Asia Pacific Neural Network Society Young Researcher Award. He received best paper or book award six times. Until October 2020, he has published 9 books and over 200 international research papers (80+ international journals) e.g., in journals (JMLR, Neural Computation, IEEE T-PAMI, IEEE T-NNLS, IEEE T-BME, IEEE T-Cybernetics) and conferences (NeurIPS, IJCAI, SIGIR, UAI, CIKM, ICDM, ICML, ECML, CVPR). He serves as associated editors/advisory board members in a number of journals and book series. He was invited as keynote speaker in more than 30 international conferences or workshops.
\end{IEEEbiography}

\begin{IEEEbiography}[{\includegraphics[width=1in,height=1in,clip,keepaspectratio]{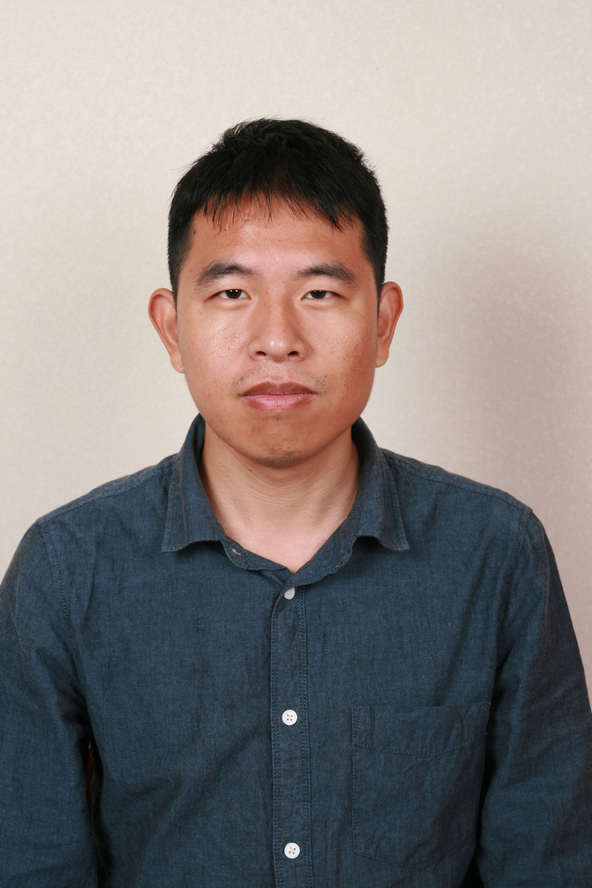}}]{Cheng Cheng}
is currently an associate professor. He received the B.S. degree and M.S. degree in Computer Science and Technology from Guizhou University, Guiyang, China, in 2004 and 2009, respectively, and the Ph.D. degree in Information Engineering from Tokyo University of Agriculture and Technology (TUAT), Japan, in 2013. His current research interests focus on 3D vision, etc.
\end{IEEEbiography}

\begin{IEEEbiography}[{\includegraphics[width=1in,height=1.25in,clip,keepaspectratio]{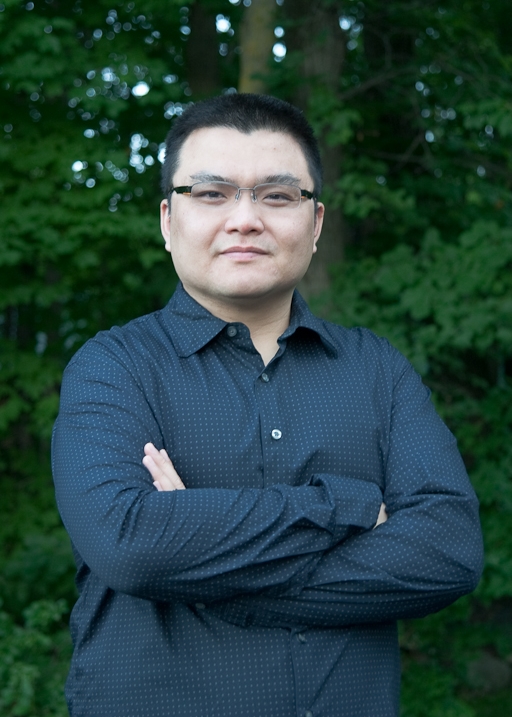}}]{Xuguang Wang}
received the Ph.D. degree from the Department of Electronic Engineering, University of Texas, Austin, USA, with a master's degree from the Department of Electronic Engineering, Rice University, USA, and a bachelor's degree from the Department of Materials, Tsinghua University, Beijing. Dr. Wang has been engaged in the research of semiconductor storage technology for 10 years, and has undertaken the research of semiconductor memory in many scientific research institutions such as National Natural Science Foundation of the United States, MARCO, SRC, etc.
\end{IEEEbiography}

\begin{IEEEbiography}[{\includegraphics[width=1in,height=1.25in,clip,keepaspectratio]{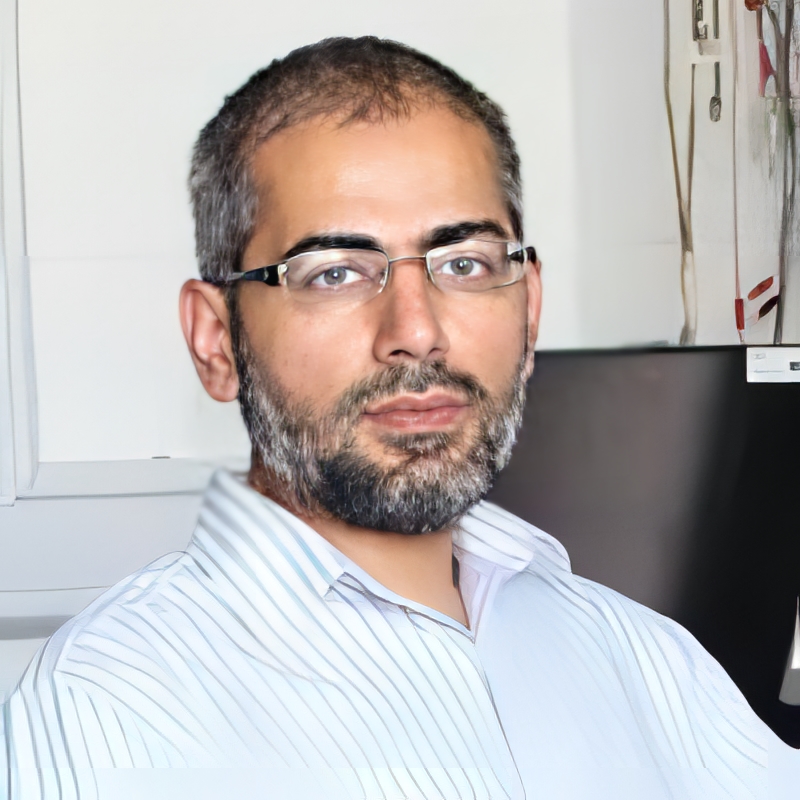}}]{Amir Hussain}
received his B.Eng (highest 1st Class Honours with distinction) and Ph.D degrees, from the University of Strathclyde, Glasgow, U.K., in 1992 and 1997, respectively. He is founding Director of the Centre of AI and Data Science at Edinburgh Napier University, UK. His research interests are cross-disciplinary and industry-led, aimed at developing cognitive data science and trustworthy AI technologies to engineer smart industrial and healthcare systems of tomorrow. He has (co)authored three international patents and around 500 publications. He is the founding Chief Editor of Springer's Cognitive Computation journal and Springer Book Series on Socio-Affective Computing. He has been invited Associate Editor/Editorial Board member for various other top journals, including the IEEE Transactions on Neural Networks and Learning Systems, Information Fusion, the IEEE Transactions on Systems, Man and Cybernetics: Systems, and the IEEE Transactions on Emerging Topics in Computational Intelligence. Amongst other distinguished roles, he is an elected Executive Committee member of the UK Computing Research Committee, General Chair of IEEE WCCI 2020, and Chair of the IEEE UK and Ireland Chapter of the IEEE Industry Applications Society.
\end{IEEEbiography}

\begin{IEEEbiography}[{\includegraphics[width=1in,height=1.25in,clip,keepaspectratio]{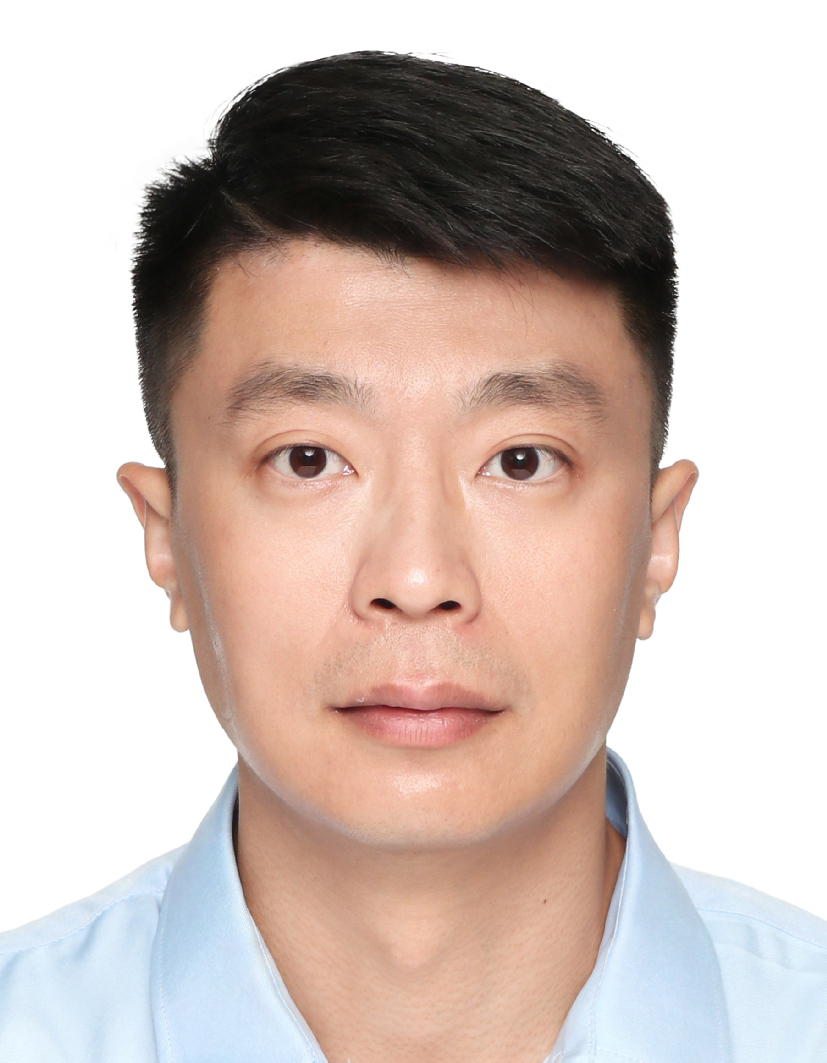}}]{Xin Liu}
(M'12) received the B.Eng. degree in electrical engineering from Tianjin University, Tianjin, China, and the Ph.D. degree in electrical engineering from Nanyang Technological University (NTU), Singapore, in 2000 and 2007, respectively. From 2007 to 2018, Dr. Liu worked as Principal Investigator and Head of Intelligent Computing Chips Department with Institute of Microelectronics, A*STAR, Singapore. Dr. Liu joined the Suzhou Institute of Nano-Tech and Nano-Bionics (SINANO), Chinese Academy of Sciences as professor in 2018. His research interests include artificial intelligence signal processing algorithms, high-performance massively parallel processing chip architecture design, ultra-low power digital processor design, embedded non-volatile memory circuit design, etc.
\end{IEEEbiography}
%


%




\end{document}